\documentclass{article}

\usepackage{arxiv}
\usepackage{authblk}

\usepackage{subcaption}
\usepackage{float}
\usepackage{comment}
\usepackage{amssymb}
\usepackage{amsmath}
\usepackage{bm}

\usepackage[utf8]{inputenc} 
\usepackage[T1]{fontenc}    
\usepackage{hyperref}       
\usepackage{url}            
\usepackage{booktabs}       
\usepackage{amsfonts}       
\usepackage{nicefrac}       
\usepackage{microtype}      
\usepackage{lipsum}
\usepackage{graphicx}
\graphicspath{ {./images/} }

\title{CoxSE: Exploring the Potential of Self-Explaining Neural Networks with Cox Proportional Hazards Model for Survival Analysis}

\author[1]{\textbf{Abdallah Alabdallah}}
\author[1]{\textbf{Omar Hamed}}
\author[1,3]{\textbf{Mattias Ohlsson}}
\author[1]{\textbf{Thorsteinn Rögnvaldsson}}
\author[1,2]{\textbf{Sepideh Pashami}}

\affil[1]{Center for Applied Intelligent Systems Research (CAISR), Halmstad University}
\affil[ ]{Halmstad, Sweden}
\affil[2]{Research Institutes of Sweden (RISE)}
\affil[ ]{Stockholm, Sweden}
\affil[3]{Centre for Environmental and Climate Science, Lund University}
\affil[ ]{Lund, Sweden}

\begin{document}
\maketitle

\begin{abstract}
The Cox Proportional Hazards (CPH) model has long been the preferred survival model for its explainability. However, to increase its predictive power beyond its linear log-risk, it was extended to utilize deep neural networks, sacrificing its explainability. In this work, we explore the potential of self-explaining neural networks (SENN) for survival analysis. We propose a new locally explainable Cox proportional hazards model, named CoxSE, by estimating a locally-linear log-hazard function using the SENN. We also propose a modification to the Neural additive (NAM) model, hybrid with SENN, named CoxSENAM, which enables the control of the stability and consistency of the generated explanations. 

Several experiments using synthetic and real datasets are presented, benchmarking CoxSE and CoxSENAM against a NAM-based model, a DeepSurv model explained with SHAP, and a linear CPH model. The results show that, unlike the NAM-based model, the SENN-based model can provide more stable and consistent explanations while maintaining the predictive power of the black-box model. The results also show that, due to their structural design, NAM-based models demonstrate better robustness to non-informative features. Among the models, the hybrid model exhibits the best robustness. Full implementation is available on GitHub.\footnote{https://github.com/abdoush/CoxSE}
\end{abstract}


\section{Introduction}
\label{introduction}

Due to their flexibility and the abundance of collected data, deep-neural-network-based models have been shown to outperform classical machine learning methods in survival modeling~\cite{ALABDALLAH2024}. However, such performance gain comes at the price of their explainability. Neural networks have long been considered black-box models because of their complicated inner workings. This makes it difficult to understand how they arrive at their predictions. Consequently, the black-box nature limits the usability of such models, especially in domains like survival analysis when used for safety-critical applications like health care or predictive maintenance~\cite{Reza:2021,Polymeri:2020,Alabdallah:2023Premature,Rahat:2023}. Nonetheless, efforts have been spent utilizing post-hoc explanation methods to compensate for the lack of transparency in black-box models~\cite{pashami:2023Xpm}.

Recently, there has been a growing interest in a type of neural network that not only predicts the outcome of an instance but also provides inherent explanations of how it arrived at such an outcome. Examples are the Neural Additive Models (NAM)~\cite{Agarwal2021} and the Self-Explaining Neural Networks (SENN)~\cite{Alvarez2018}. Such models can learn to fit the problem while providing intrinsic explanations for their decisions. 
The NAM method converts the problem into a sum of single-input functions of individual features, modeled as a neural network. The output of these functions represents the respective feature's contribution to the decision. The SENN models the function as a locally-linear function where the model learns the relevance of the features conditioned on the input.

The NAM structure was first used for survival analysis in SurvNAM~\cite{Utkin2022} as a surrogate Cox-based model to explain black-box survival models. Subsequently, it was recently proposed as a stand-alone model named CoxNAM~\cite{Xu2023}. However, owing to their structural design, NAM models are unable to capture interactions between features, thereby constraining their predictive capacity. Moreover, the NAM models provide explanations that highlight the contribution of each feature to the decision, omitting the model's sensitivity to changes in those features.
In contrast, SENN-based models are not constrained by limitations in modeling feature interactions and can attain predictive performance comparable to that of black-box models. Additionally, these models learn the relevance of individual features, reflecting their local sensitivities, from which the contributions of features to specific predictions can be derived. 

In this work, we explore the potential of the SENN structure for survival modeling. We propose CoxSE, a Cox-based self-explaining model relying on the SENN structure. CoxSE learns features' relevance, which can be interpreted as the weights of the line tangent to the decision boundary in the vicinity of the point of interest. As all features contribute to the computation of each feature's relevance, CoxSE can account for feature interactions. Additionally, CoxSE has two extra regularization terms to encourage the stability and consistency of the provided explanations.

However, such flexibility comes with a price, as we demonstrate that the NAM structure is more robust to noise. Therefore, we also propose a hybrid model, CoxSENAM, adopting the NAM structure with the SENN type of output and loss function. Similar to CoxNAM, this model is more suitable for cases with no feature interactions, but benefits from the extra regularization of the learned features' relevance, achieving better stability, consistency, and robustness to noise.

In summary, this paper provides three contributions: 1) proposing CoxSE, a new Cox-based survival model providing intrinsic explanations while maintaining the similar performance of its black-box counterpart; 2) proposing CoxSENAM, an enhancement to the NAM-based model adopting the SENN type of output and loss function which shows better explanation stability, consistency, and robustness to noise; 3) illustrating the pros and cons of the proposed and existing models.

\section{Survival Analysis}
Survival analysis is a branch of statistics concerned with studying the time until an event of interest occurs—for example, a patient's death or a machine's failure. The main challenge that gave rise to survival analysis is that survival data contains subjects who have not experienced the event yet at the end of the study. Such cases are called censored cases, which are assumed to experience the event at a later time greater than their recorded time, in which case it is called right-censored data. Other types of censoring exist, but right censoring is the most common. 

Survival data usually comes as tuples of three variables ($\bm{x}_i, t_i, e_i$) that describe, for each subject ($i$), the feature vector ($\bm{x}_i$), the recorded time ($t_i$), and the event indicator ($e_i$) to indicate whether the recorded time is the time-to-event or the censoring time.

The survival and hazard functions are the two main functions that survival analysis usually endeavors to estimate. Given survival time as a random variable $T$, the survival function is defined as the probability of survival beyond time $t$:
\begin{equation}
    S(t) = P(T > t)
    \label{eq:survival_finction}
\end{equation}
The hazard function represents the instantaneous death rate, which quantifies the risk of the event in the infinitesimal time $\Delta t$ given that the subject has survived until the time $t$ and is defined as
\begin{equation}
    h(t) = \lim_{\Delta t \to 0} \frac{P(t \leq T < t+\Delta t|T \geq t)}{\Delta t}
    \label{eq:hazard_function}
\end{equation}
The two functions are related through the relation
\begin{equation}
    h(t) = - \frac{d}{dt} \ln (S(t))
\end{equation}

The cumulative hazard function (CHF) is another function of interest to some models and is related to the hazard and survival functions by
\begin{equation}
   H(t) = \int_{0}^{t} h(\tau) d\tau 
\end{equation}
and
\begin{equation}
    S(t) = e^{-H(t)}
\end{equation}

Early on, statistical methods were developed to estimate such functions as the Kaplan-Meier estimator~\cite{Kaplan:1958} for the survival function, and the Nelson-Aalen estimator~\cite{Nelson:1969,Nelson:1972,aalan:1978} for the CHF. However, these methods do not account for feature dependencies. The Cox Proportional Hazards Model (CPH)~\cite{Cox1972} was the first statistical model to incorporate feature dependence in the analysis of time-to-event data. The CPH model assumes a baseline hazard function $h_0(t)$, which is common to all subjects and solely dependent on time, while the covariates $\bm{x}$ exert time-independent, multiplicative effects on the hazard through an exponential function. Consequently, the model imposes a linear relationship between the covariates and the logarithm of the hazard function: 
\begin{equation}
    h(t, \bm{x}) = h_0(t) e^{\bm{w}^{\intercal} \bm{x}}
    \label{eq:cph_ht}
\end{equation}
Due to its linearity, the CPH model is regarded as inherently explainable, with the weights $\bm{w}$ representing the relative importance of the features in influencing the hazard.

Over time, machine learning techniques have been increasingly applied to survival analysis, enabling the development of more accurate and flexible models. More recently, the widespread availability of large-scale data has led to the emergence of deep learning models as the dominant approach in this domain. Numerous deep-learning models have been proposed to estimate different survival analysis functions, employing various techniques and objective functions for this purpose~\cite{Katzman:2018,Lee_2018,ALABDALLAH2024,Altarabichi:2024Loss}. Interestingly, more than $40\%$ of such models employ the Cox-based formulation~\cite{Wiegrebe2024}. Most notably, DeepSurv~\cite{Katzman:2018} which extends the CPH linear log-risk function ($\bm{w}^{\intercal} \bm{x}$ in Eq.~\ref{eq:cph_ht}) to a non-linear function ($f(\bm{x})$ in Eq.~\ref{eq:deepsurv_ht}) modeled as a neural network. 
\begin{equation}
    h(t, \bm{x}) = h_0(t) e^{f(\bm{x})}
    \label{eq:deepsurv_ht}
\end{equation}

With the growing variety and complexity of survival models, robust and consistent evaluation becomes essential. Among the many metrics available, the concordance index (C-index)~\cite{Harrell:1982} is arguably the most widely used. The C-index (\emph{CI}) quantifies the rank correlation between the predicted scores of subjects and their observed or censored times. More specifically, it represents the probability that the predicted risk scores of two randomly selected subjects agree with the order of their observed event times:
\begin{equation}
    \mbox{\emph{CI}} = \mathbb{P}(\hat{r}_i > \hat{r}_j | t_i < t_j, e_i=1)
    \label{eq:c_index}
\end{equation}
where $\hat{r}_{i}$ is the predicted risk score of the subject $i$ with the time-to-event value $t_i$ and event indicator $e_i$. A good model should assign higher risk scores for subjects with shorter survival times, leading to a higher C-index value.

The concordance index is closely related to the CPH model in the sense that both focus on the rank order of events and are insensitive to any monotonic transformation. Such a relation makes the C-index suitable for our purpose as all the models in this work follow the assumptions of the Cox Proportional Hazards model.

\section{Related work}
\subsection{Post-Hoc Explainability Methods}
As machine learning started to be used for survival analysis, a need for suitable explanation methods emerged. Many post-hoc explanation methods in machine learning like LIME~\cite{Ribeiro_2016} and SHAP~\cite{Lundberg_2017} have been adapted to survival analysis models as in~\cite{KOVALEV2020,Alabdallah2022,KRZYZINSKI2023,Langbein_2025,Baniecki_2025}. Moreover, other methods relied on counterfactual examples, adapting them for survival models~\cite{Kovalev:2021Counterfactuals,Alabdallah:2024Counterfactuals}. 

\subsection{Inherently Explainable Models}
A growing research direction focuses on building inherently explainable, yet expressive, models. SurvNAM~\cite{Utkin2022} proposed the use of Neural Additive Models (NAM)~\cite{Agarwal2021} and an extended Cox-Proportional Hazards model as a surrogate model to explain black-box survival models. More recently, CoxNAM~\cite{Xu2023} proposed using the NAM structure with the Cox formulation as a standalone inherently explainable survival model. 

Originally, the NAM models the relation between an input $\bm{x}=(x_0, x_1, ..., x_p)$ and a target $y$ as an additive function based on the Generalized Additive Models (GAM) model~\cite{Hastie1986}:
\begin{equation}
    \mathbb{E}(y)= \sum_{i=0}^{p} g_{i}(x_i)
    \label{eq:nam}
\end{equation}
where in the NAM model, the $g_{i}$ functions are modeled as neural networks. 

\begin{figure}[H]
    \centering
    \includegraphics[width=0.7\columnwidth]{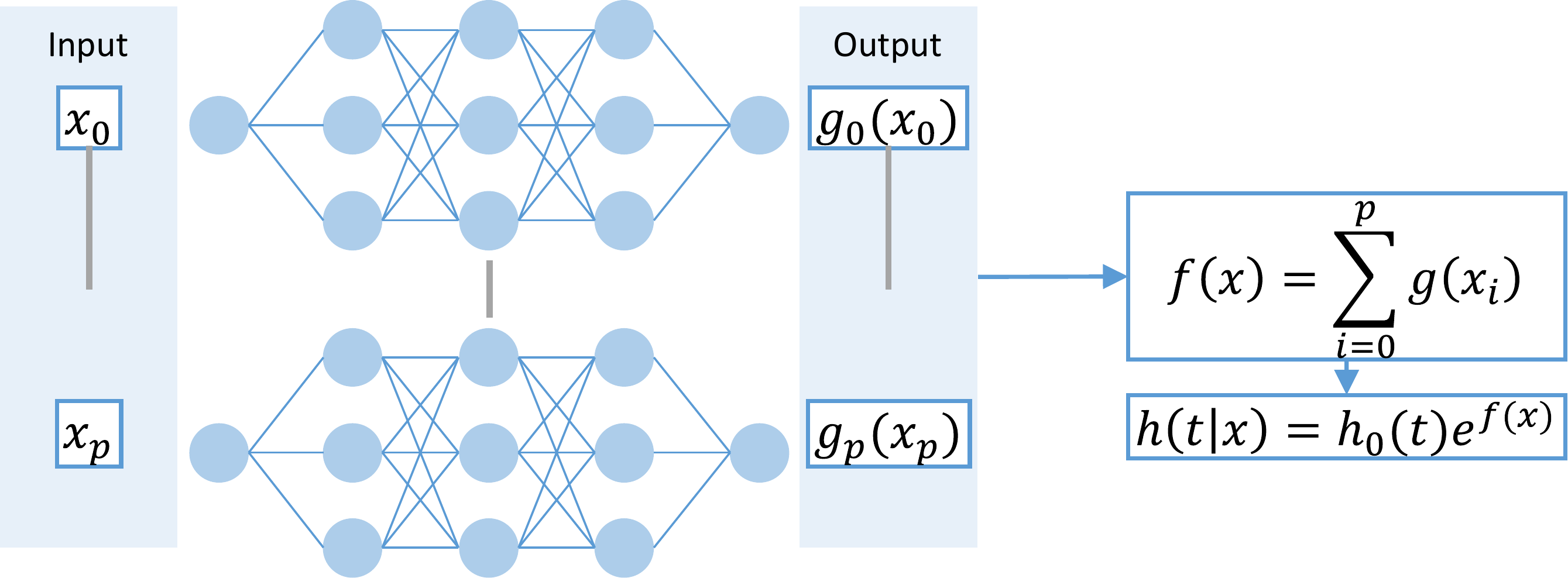}
    \caption{CoxNAM Model architecture}
    \label{fig:CoxNAM}
\end{figure}

The CoxNAM model uses the same scheme, however, to model the hazard function based on the Cox Proportional Hazards formulation as illustrated in Figure~\ref{fig:CoxNAM}.
Due to their design, NAM-based models cannot model interactions between features. This limits the expressive power of such models, leading to underfitting in most cases. Moreover, there is no guarantee that the model will learn valid explanations.

\begin{figure}[H]
    \centering
    \includegraphics[width=0.6\columnwidth]{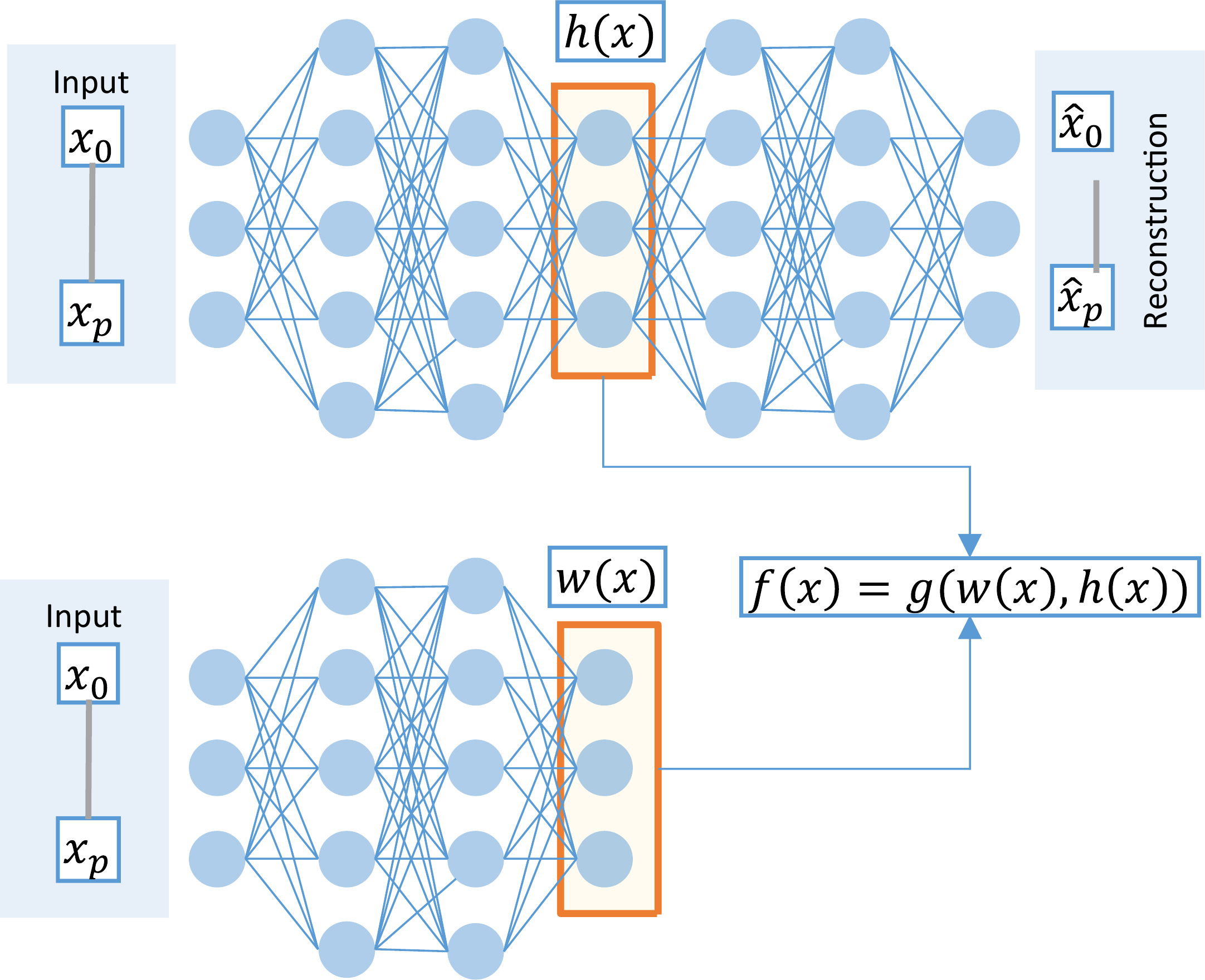}
    \caption{SENN Model architecture}
    \label{fig:SENN}
\end{figure}

Our work builds on another interesting direction for inherent explanations: the Self-Explaining Neural Networks (SENN)~\cite{Alvarez2018}. Such a model has two branches: a concept encoder that generates interpretable features $\bm{h}(\bm{x})$ from the input and an input-dependent parameterizer that learns the features' relevance scores (weights) $\bm{w}(\bm{x})$, as illustrated in Figure~\ref{fig:SENN}. 

Eventually, the function can be written as:
\begin{equation}
    f(x)= g(w_1(\bm{x})h_1(\bm{x}),..,w_p(\bm{x})h_p(\bm{x})) 
\end{equation}
%
%
where $p$ is the number of features and $g(.)$ is the aggregation function, which can be any affine function with positive weights or simply an addition, approximating the function with a locally-linear function as:
\begin{equation}
    f(x)= \sum_{x_i \in \bm{x}} w_i(\bm{x})h_i(\bm{x})
    \label{eqn:senn_add}
\end{equation}
Additionally, SENNs regularize their generated explanations to ensure that similar subjects have similar explanations, a quality referred to as stability.

For tabular data, which is the case in our work, the concept vector is the original feature space, i.e. $\bm{h}(\bm{x})=\bm{x}$. This eliminates the need for the autoencoder branch of the SENN structure, along with its reconstruction part of the loss function. 

As the weights of each feature depend on all the features, SENN can model the interactions between the features, maintaining similar performance as regular neural networks. Moreover, SENN, with its regularization, can control the quality of the explanations.

\section{Method}
In this work, we investigate the use of Self-Explaining Neural Networks (SENN) and Neural Additive Models (NAM) to improve interpretability in survival analysis. We adapt the SENN framework to the Cox Proportional Hazards (CPH) model and integrate key elements from both SENN and NAM to develop two novel models: CoxSE and CoxSENAM. Our aim is to provide locally interpretable risk predictions while retaining the flexibility of non-linear models.

\subsection{Self-Explaining Cox Model (CoxSE)}
The CoxSE method extends SENN to survival analysis using the CPH model, but with a non-linear relationship $f(\bm{x})$ between the predictors and the hazard:

\begin{equation}
    h(t, \bm{x}) = h_0(t) e^{f(\bm{x})}
    \label{eq:coxse_ht}
\end{equation}

where $h_0(t)$ is an unspecified baseline hazard function and the log-risk function $f(\bm{x})$ is approximated as a locally linear function
\begin{equation}
    f(\bm{x}) = \sum_{x_i \in \bm{x}} w_{\theta_i}(\bm{x}) x_i
    \label{eq:coxse}
\end{equation}

\begin{figure}[H]
\centering
\begin{subfigure}{0.50\columnwidth}
  \centering
  \includegraphics[width=0.8\linewidth]{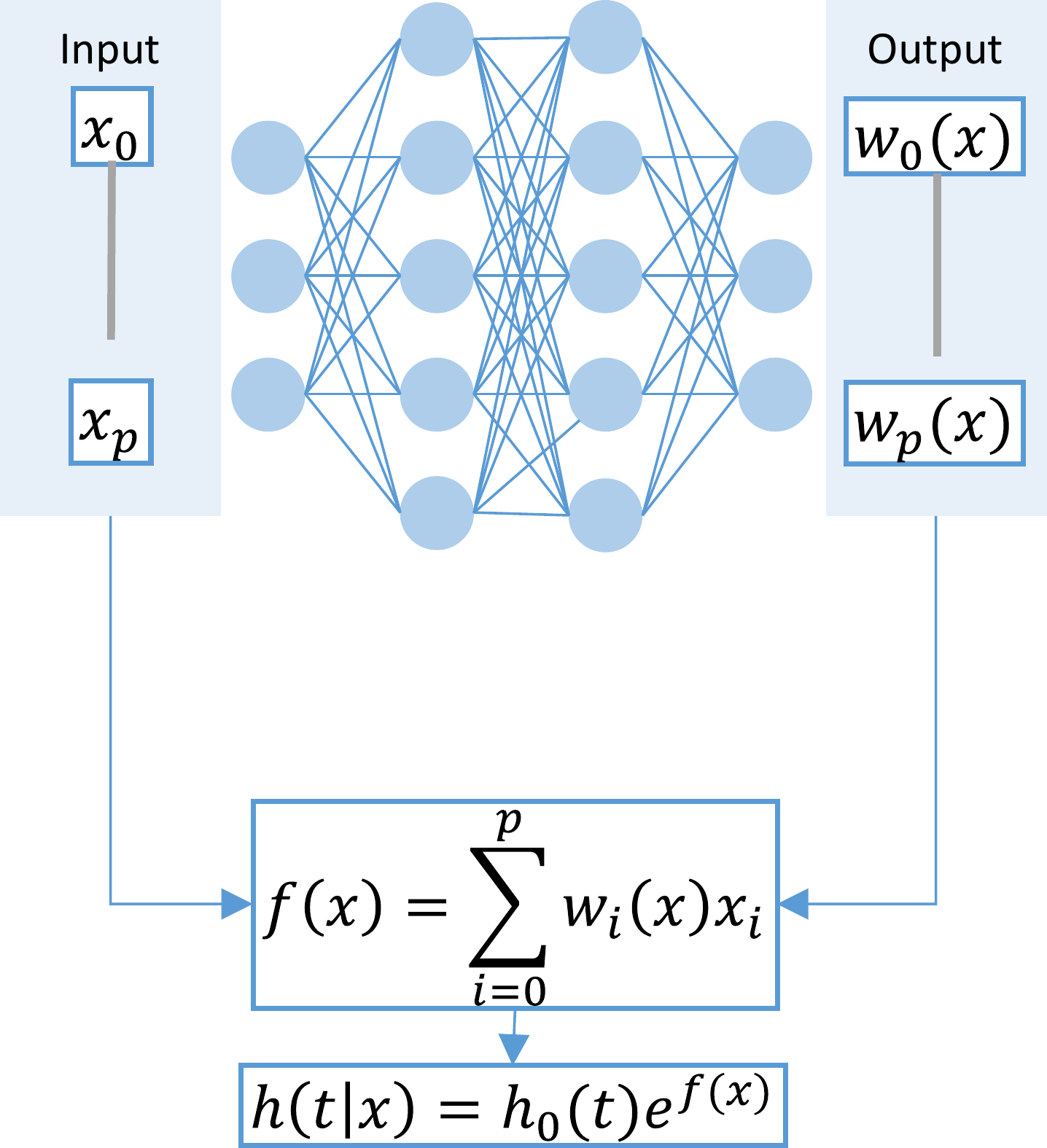}
  \caption{CoxSE}
  \label{fig:CoxSE}
\end{subfigure}%
~
\begin{subfigure}{0.50\columnwidth}
  \centering
  \includegraphics[width=0.8\linewidth]{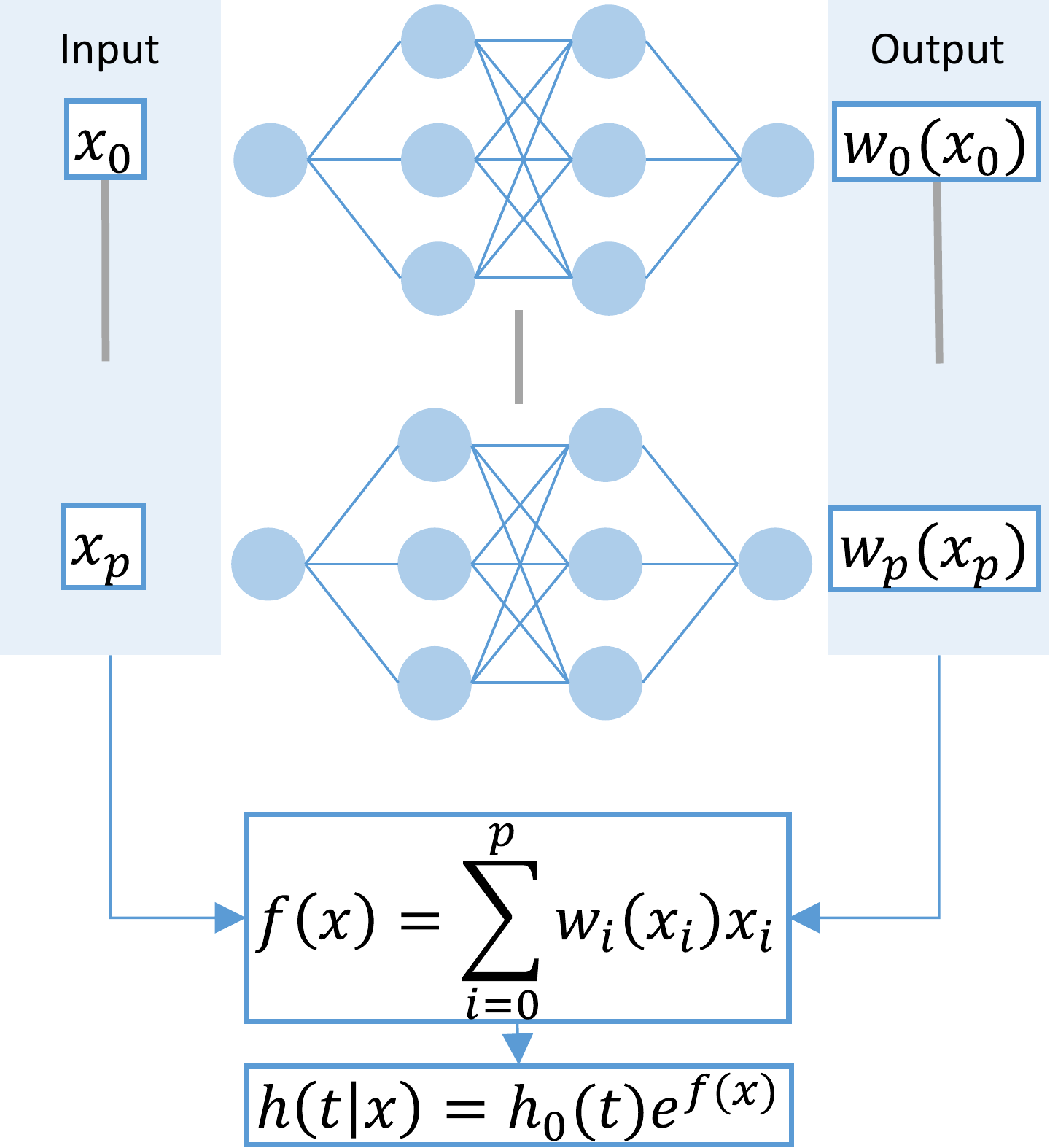}
  \caption{CoxSENAM}
  \label{fig:CoxSENAM}
\end{subfigure}
\caption{CoxSE and CoxSENAM models' architecture}
\label{fig:CoxSE_CoxSENAM}
\end{figure}

This allows the model to learn the relevance vector $\bm{w}_{\theta}(\bm{x})$ that specifies the linear weights at each point $\bm{x}$. Such weights describe a vector perpendicular to the decision boundary and reflect the local features' importance at each specific location. 
Our model deals with survival data, which mostly comes as tabular data representing subjects' features. Such features are usually explainable by themselves. This eliminates the role of the autoencoder part of the SENN model in extracting interpretable representation $\bm{h}(\bm{x})$, reducing it to the original feature space, i.e., $\bm{h}(\bm{x})=\bm{x}$, which is reflected in our model's function shown in Eq.~\ref{eq:coxse}. 
An illustration of the CoxSE model is shown in Figure~\ref{fig:CoxSE}.

\subsection{CoxSENAM: A Hybrid SENN + NAM Model}
Based on the SENN and the NAM models, we also propose a hybrid model named CoxSENAM. As illustrated in Figure~\ref{fig:CoxSENAM}, the model has a structure similar to the CoxNAM model and a type of output and loss function similar to the CoxSE model. 

The only difference between CoxSE and CoxSENAM is that the feature relevances are computed independently:
\begin{equation}
    f(\bm{x}) = \sum_{x_i \in \bm{x}} w_{\theta_i}(x_i) x_i
    \label{eq:coxsenam}
\end{equation}
where $w_{\theta_i}$ is the relevance function of the i'th feature ($x_i$) in the feature vector $\bm{x}$.
The independence in computing feature relevances, although it constrains the model’s capacity to capture complex feature interactions, can contribute to mitigating the impact of noisy features—a benefit that will be demonstrated in subsequent analyses. On the other hand, this hybrid formulation enables the NAM model to leverage both regularization terms in the loss function, detailed in the following section, enhancing its stability and robustness. 

\subsection{Loss Function}
The loss function for both CoxSE and CoxSENAM is composed of three terms. The first term is the Partial Likelihood function inherited from the CPH model (Eq.~\ref{eqn_partial_likelihood}). This part of the loss function maximizes the concordance between the predicted hazards.
\begin{equation}
    \mathcal{L}_1(\theta) = \prod_{i=1}^{N} \frac{e^{f_{\theta}(\bm{x}_i)}}{\sum_{\bm{x}_j \in R_i} e^{f_{\theta}(\bm{x}_j)}}
    \label{eqn_partial_likelihood}
\end{equation}
where $R_i$ are the subjects at risk at time $t_i$.

The second term (Eq.~\ref{eqn_w_loss2}) minimizes the distance between $\nabla_{x} f_{\theta}(\bm{x})$, the gradient of $f_{\theta}(\bm{x})$ with respect to $\bm{x}$, and the predicted local weights $w_{\theta}(\bm{x})$:
\begin{equation}
    \mathcal{L}_2(\theta) = \|\nabla_{x} f_{\theta}(\bm{x}) - w_{\theta}(\bm{x}) \|_2
    \label{eqn_w_loss2}
\end{equation}
This term is adopted from the SENN model to encourage explanation stability.
A third term is added to regularize the learned weights and encourage the model to select a subset of the most important features while setting the weights of less important features to zero:
\begin{equation}
    \mathcal{L}_3(\theta) = \|w_{\theta}(\bm{x}) \|_1
    \label{eqn_w_loss3}
\end{equation}
The three terms comprise the complete loss function:
%
\begin{equation}
    \mathcal{L}(\theta) = \mathcal{L}_1(\theta) + \alpha \mathcal{L}_2(\theta) + \beta \mathcal{L}_3(\theta)
    \label{eqn_CoxSE_loss}
\end{equation}

\section{Experiments}
\subsection{Datasets}

Three synthetic datasets are created with different characteristics based on the proportional hazards assumption following the simulations in~\cite{Kvamme:2019}. Moreover, two real datasets were used in the experiments. 

\textbf{Lin Dataset:}
A synthetic dataset that consists of $50,000$ subjects with three features of increasing importance, and a linear relationship between the log-risk $f(\bm{x})$ and the predictors as follows:
\begin{equation}
    f(\bm{x}) = 0.44 x_0 + 0.66 x_1 + 0.88 x_2
    \label{eq:lin_dataset}
\end{equation}

\textbf{NonLin Dataset:}
A synthetic dataset that consists of $50,000$ subjects with three features and a non-linear relationship between the log-risk $f(\bm{x})$ and the predictors, with different importance represented by the feature weights as follows:
\begin{equation}
    f(\bm{x}) = 3 x_0^2 + 2 x_1^2 + x_2^2
    \label{eq:nonlin_dataset}
\end{equation}

\textbf{NonLinX Dataset:}
A synthetic dataset that consists of $50,000$ subjects with three features and a non-linear relationship between the log-risk $f(\bm{x})$ and the predictors, different importance represented by the feature weights, and an interaction between the first two features, as follows:
\begin{equation}
    f(\bm{x}) = 3 x_0^2 + 2 x_1^2 + x_2^2 + 3 x_0 x_1
    \label{eq:nonlinx_dataset}
\end{equation}

\textbf{FLCHAIN Dataset:}
This survival analysis benchmark dataset is used to study whether the free light chain (FLC) assay is a good predictor of survival time~\cite{Dispenzieri:2012}. The dataset consists of $7,874$ subjects with $27$ numerical and one-hot-encoded features, where the FLC is the sum of the two features \texttt{kappa} and \texttt{lambda}. As a result, the study found that a high FLC value was a significant predictor of worse survival.

\textbf{SEER Dataset:}
The Surveillance, Epidemiology, and End Results (SEER) database~\cite{seer:2016}, maintained by the US National Cancer Institute, is a comprehensive dataset that gathers detailed information on cancer patients. This dataset includes data on diagnoses, treatments, and survival rates. We utilized version 8.4.3 of the "SEER*Stat" software and the "Incidence – SEER research data, 17 registries" release from November 2022. The dataset encompasses cancer incidences from 2000 to 2020. However, we narrowed our analysis to incidences between 2004 and 2015 to avoid preprocessing challenges due to different coding systems used across various years. The dataset consists of $458,117$ subjects with $32$ numerical, label-encoded, and one-hot-encoded features.

\subsection{Experiment Settings}

Five models, all based on Cox Proportional Hazards, are compared: CPH, DeepSurv, CoxNAM, CoxSE, and CoxSENAM. All those models are self-explaining except for DeepSurv, for which the SHAP method is used to explain. Experiments on the three synthetic and the two real datasets were conducted. To provide benchmarking context, we also report results from two non-Cox approaches, Random Survival Forest (RSF)~\cite{Ishwaran_2008} and DeepHit~\cite{Lee_2018}. However, as our focus is on interpretability within Cox-based frameworks, these baselines are not included in the detailed discussion. Five-fold cross-validation is used to tune the models. More details about the hyperparameters are provided in~\ref{appendix_a}. 

\subsection{Performance of the Models}
The results in Table~\ref{tbl_c_index_all} show that all models performed similarly on the Lin dataset. Also, the four deep-learning-based models performed similarly on the NonLin dataset except for the CPH model, which could not model the non-linear relationship in the data. The more remarkable case is the NonLinX, the non-linear synthetic dataset with interactions between the features. On this dataset, both DeepSurv and CoxSE performed similarly. However, CoxNAM and CoxSENAM failed to model the feature interaction and converged to a significantly lower C-index. In this case, CPH also failed to model the data and converged to a random model. This difference in performance on the NonLinX dataset is due to the structural design of the NAM-based models, CoxNAM and CoxSENAM, which cannot handle feature interactions.

In the cases of the real datasets, all models except CPH performed similarly on the FLCHAIN dataset. Intriguingly, the models show three levels of performance on the SEER dataset. The best performance is demonstrated by CoxSE and DeepSurv, with no significant difference between them. On the other hand, CoxNAM and CoxSENAM also performed similarly, but significantly below CoxSE and DeepSurv. Finally, CPH performed significantly lower than the four other models. Such results on the SEER dataset could indicate non-linearity and interaction between the features.

\begin{table*}[ht]
\caption{Concordance Index (C-Index) performance of all models across all datasets. Values in parentheses (.) denote statistically significant differences based on a paired t-test over 5-fold cross-validation.}
\centering
\resizebox{\textwidth}{!}{%
\begin{tabular}{l c c c c c} 
 \hline
 Model/Dataset & Lin & NonLin & NonLinX & FLCHAIN & SEER \\ 
 \hline
CPH      & 67.67$\pm$0.020 & (50.07$\pm$0.078) & (50.01$\pm$0.069) & (78.33$\pm$0.094) & (80.05$\pm$0.103) \\
DeepSurv & 67.68$\pm$0.001 & 75.30$\pm$0.011   & 77.92$\pm$0.011   & 78.45$\pm$0.095   & 84.35$\pm$0.027   \\
CoxNAM   & 67.67$\pm$0.011 & 75.30$\pm$0.012   & (70.84$\pm$0.092) & 78.59$\pm$0.169   & (83.86$\pm$0.018) \\
\hline
CoxSE (ours)    & 67.68$\pm$0.001 & 75.29$\pm$0.024   & 77.93$\pm$0.021   & 78.57$\pm$0.031   & 84.37$\pm$0.018   \\ 
CoxSENAM (ours) & 67.67$\pm$0.012 & 75.31$\pm$0.006   & (70.93$\pm$0.118) & 78.51$\pm$0.057   & (83.85$\pm$0.029) \\
\hline
\multicolumn{6}{c}{Reference baselines (non-Cox) models} \\
\hline
RSF      & (67.44$\pm$0.032) & (74.45$\pm$0.056) & (76.62$\pm$0.054) & (78.39$\pm$0.096) & (84.10$\pm$0.011) \\
DeepHit  & 67.67$\pm$0.004 & (75.21$\pm$0.017)   & (77.88$\pm$0.032)   & 78.58$\pm$0.071   & (84.28$\pm$0.041)   \\
\hline
\end{tabular}
}

\label{tbl_c_index_all}
\centering
\end{table*}

Building upon the previous analysis of model performance, we further report the average execution time per run for each model in Table~\ref{tbl_time}. The results indicate that CoxSE, CoxSENAM, CoxNAM, and DeepSurv exhibit similar execution times across the majority of datasets. Notably, more substantial differences were observed on the SEER dataset—the largest in the study—where CoxSENAM required the longest time to complete model fitting.

\begin{table*}[ht]
\caption{Average execution time (in seconds) of the models' 5-fold runs on each dataset.}
\centering
\resizebox{\textwidth}{!}{%
\begin{tabular}{l c c c c c c} 
 \hline
Model/Dataset & Lin & NonLin & NonLinX & FLCHIAN & SEER & Avg.$\pm$std.\\ 
\hline
CPH & 16.0 & 26.1 & 12.8 & 8.3 & 151.1 & 42.9$\pm$60.9 \\ 
DeepSurv & 17.7 & 56.6 & 181.9 & 27.6 & 490.9 & 155.0$\pm$198.9 \\ 
CoxNAM & 16.3 & 23.0 & 26.1 & 32.1 & 1285.0 & 276.5$\pm$563.8 \\ 
\hline
CoxSE (ours) & 20.5 & 123.4 & 21.4 & 18.2 & 908.6 & 218.4$\pm$388.4 \\ 
CoxSENAM (ours) & 18.0 & 31.5 & 22.3 & 123.8 & 3128.9 & 664.9$\pm$1378.1 \\ 
\hline
\multicolumn{7}{c}{Reference baselines (non-Cox) models} \\
\hline
RSF & 407.3 & 439.6 & 428.7 & 1.2 & 59.0 & 267.2$\pm$217.7 \\ 
DeepHit & 335.0 & 472.9 & 1906.1 & 106.7 & 3553.3 & 1274.8$\pm$1456.1 \\ 
\hline
\end{tabular}
}
\label{tbl_time}
\centering
\end{table*}

\subsection{Aggregated Explanations}
The aggregated explanations are computed as the mean of the absolute values of the explanations of the explained subjects:
\begin{equation}
    F_{expl}^i = \frac{1}{n} \sum_{j=0}^{n-1} |f_{expl}^i(\bm{x}_j)|
\end{equation}
where $f_{expl}^i(.)$ is the local explanation value returned by the model of the feature of index $i$, $\bm{x}_j$ is the feature vector of the subject $j$, and $n$ is the number of explained subjects. In particular, these local explanations are $w_i$ for the CPH model (Eq.~\ref{eq:cph_ht}), $w_{\theta_i}(\bm{x})$ for CoxSE (Eq.~\ref{eq:coxse}), $w_{\theta_i}(x_i)$ for CoxSENAM (Eq.~\ref{eq:coxsenam}), $g_{i}(x_i)$ for CoxNAM (Eq.~\ref{eq:nam}), and Shapley values for DeepSurv.

\begin{figure*}[ht]
\centering
\begin{subfigure}{0.20\linewidth}
  \centering
  \includegraphics[width=1\linewidth]{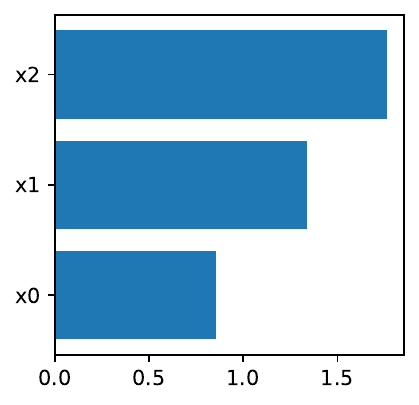}
  \caption{CoxSE}
  \label{fig:SimStudyLinearPH_CoxSE_CoxSE_W}
\end{subfigure}%
~
\begin{subfigure}{0.20\linewidth}
  \centering
  \includegraphics[width=1\linewidth]{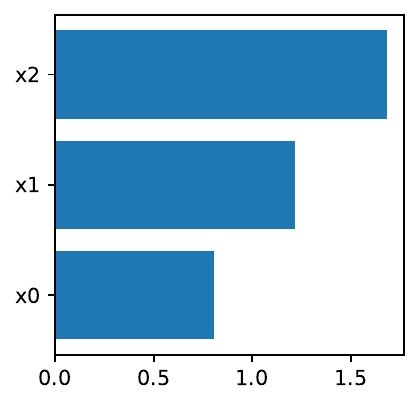}
  \caption{CoxSENAM}
  \label{fig:SimStudyLinearPH_CoxSENAM_CoxSENAM_W}
\end{subfigure}%
~
\begin{subfigure}{0.20\linewidth}
  \centering
  \includegraphics[width=1\linewidth]{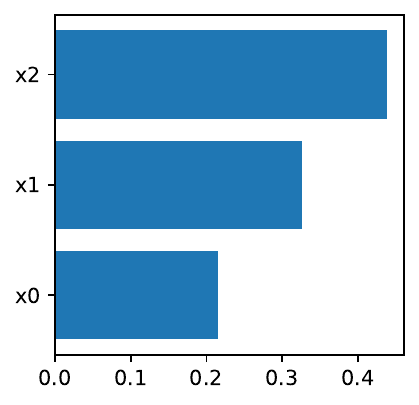}
  \caption{DeepSurv/SHAP}
  \label{fig:SimStudyLinearPH_DeepSurv_DeepSurv_WX}
\end{subfigure}%
~
\begin{subfigure}{0.20\linewidth}
  \centering
  \includegraphics[width=1\linewidth]{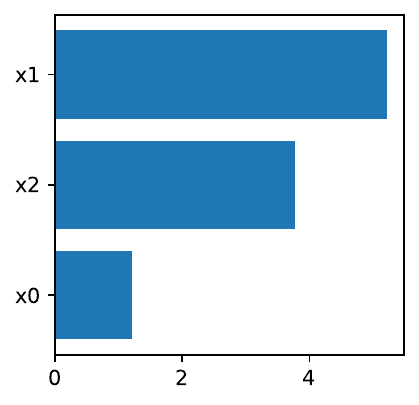}
  \caption{CoxNAM}
  \label{fig:SimStudyLinearPH_CoxNAM_CoxNAM_WX}
\end{subfigure}%
~
\begin{subfigure}{0.20\linewidth}
  \centering
  \includegraphics[width=1\linewidth]{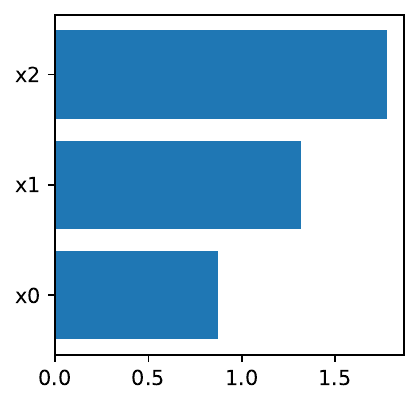}
  \caption{CPH}
  \label{fig:SimStudyLinearPH_CPH_CPH_W}
\end{subfigure}
\caption{Lin Dataset Aggregated Explanations}
\label{fig:SimStudyLinearPH_explainations}
\end{figure*}

\begin{figure*}[ht]
\centering
\begin{subfigure}{0.20\linewidth}
  \centering
  \includegraphics[width=1\linewidth]{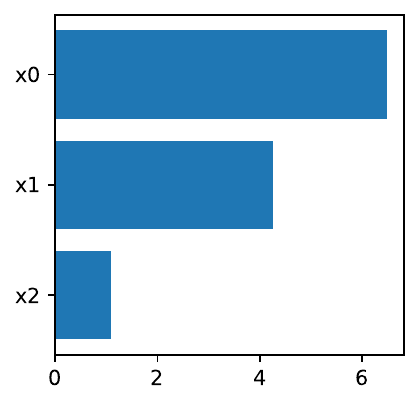}
  \caption{CoxSE}
  \label{fig:SimNonLinearPH3_0_CoxSE_CoxSE_W}
\end{subfigure}%
~
\begin{subfigure}{0.20\linewidth}
  \centering
  \includegraphics[width=1\linewidth]{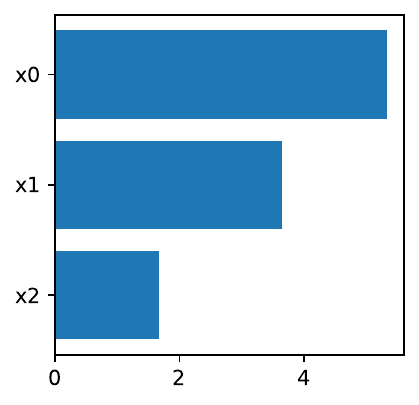}
  \caption{CoxSENAM}
  \label{fig:SimNonLinearPH3_0_CoxSENAM_CoxSENAM_W}
\end{subfigure}%
~
\begin{subfigure}{0.20\linewidth}
  \centering
  \includegraphics[width=1\linewidth]{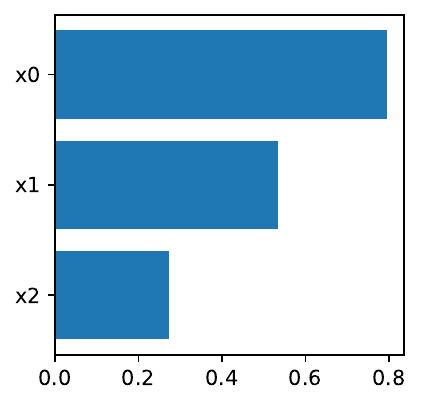}
  \caption{DeepSurv/SHAP}
  \label{fig:SimNonLinearPH3_0_DeepSurv_DeepSurv_WX}
\end{subfigure}%
~
\begin{subfigure}{0.20\linewidth}
  \centering
  \includegraphics[width=1\linewidth]{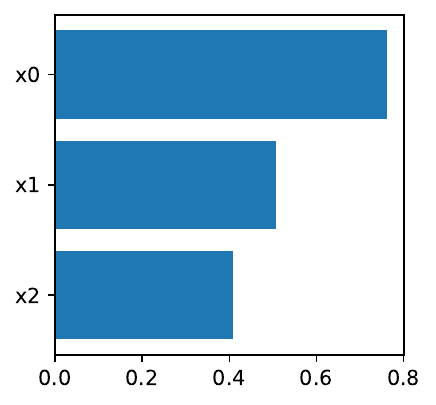}
  \caption{CoxNAM}
  \label{fig:SimNonLinearPH3_0_CoxNAM_CoxNAM_WX}
\end{subfigure}%
~
\begin{subfigure}{0.20\linewidth}
  \centering
  \includegraphics[width=1\linewidth]{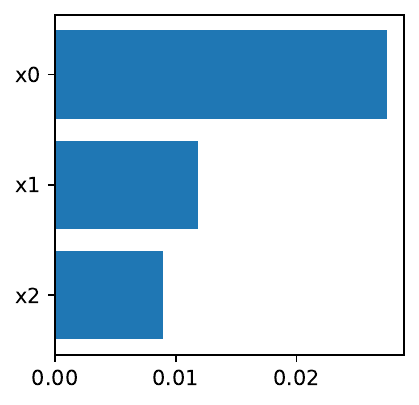}
  \caption{CPH}
  \label{fig:SimNonLinearPH3_0_CPH_CPH_W}
\end{subfigure}
\caption{NonLin Dataset Aggregated Explanations}
\label{fig:SimNonLinearPH3_0_explainations}
\end{figure*}

\begin{figure*}[ht]
\centering
\begin{subfigure}{0.20\linewidth}
  \centering
  \includegraphics[width=1\linewidth]{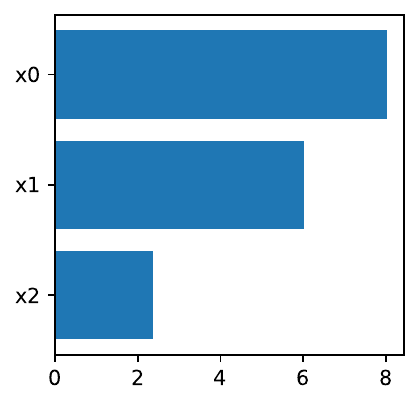}
  \caption{CoxSE}
  \label{fig:SimNonLinearPH3_3_CoxSE_CoxSE_W}
\end{subfigure}%
~
\begin{subfigure}{0.20\linewidth}
  \centering
  \includegraphics[width=1\linewidth]{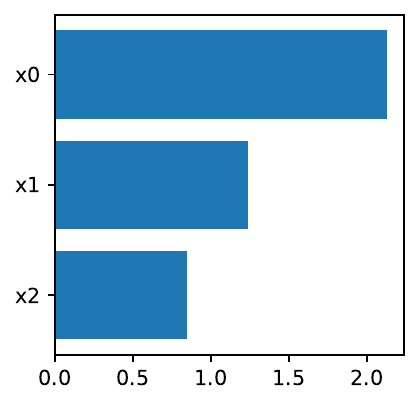}
  \caption{CoxSENAM}
  \label{fig:SimNonLinearPH3_3_CoxSENAM_CoxSENAM_W}
\end{subfigure}%
~
\begin{subfigure}{0.20\linewidth}
  \centering
  \includegraphics[width=1\linewidth]{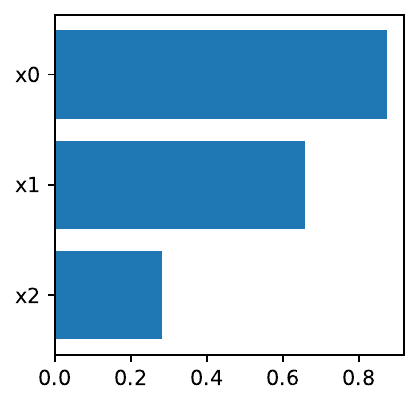}
  \caption{DeepSurv/SHAP}
  \label{fig:SimNonLinearPH3_3_DeepSurv_DeepSurv_WX}
\end{subfigure}%
~
\begin{subfigure}{0.20\linewidth}
  \centering
  \includegraphics[width=1\linewidth]{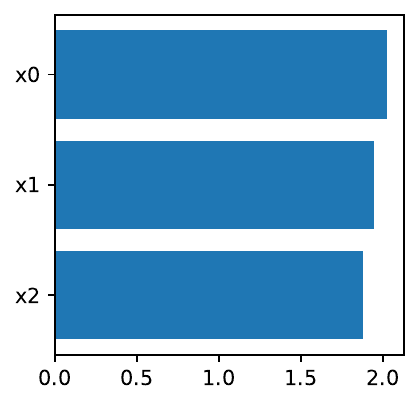}
  \caption{CoxNAM}
  \label{fig:SimNonLinearPH3_3_CoxNAM_CoxNAM_WX}
\end{subfigure}%
~
\begin{subfigure}{0.20\linewidth}
  \centering
  \includegraphics[width=1\linewidth]{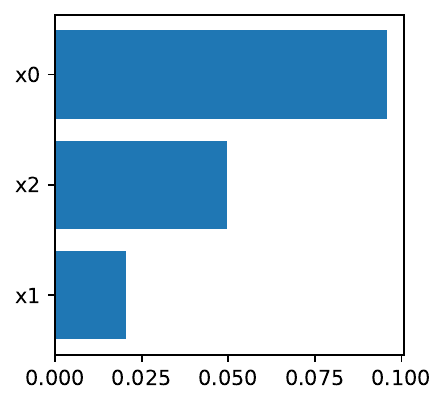}
  \caption{CPH}
  \label{fig:SimNonLinearPH3_3_CPH_CPH_W}
\end{subfigure}
\caption{NonLinX Dataset Aggregated Explanations}
\label{fig:SimNonLinearPH3_3_explainations}
\end{figure*}

The results illustrated in Figures~\ref{fig:SimStudyLinearPH_explainations},~\ref{fig:SimNonLinearPH3_0_explainations},~and~\ref{fig:SimNonLinearPH3_3_explainations} show the aggregated explanations of the models on the three synthetic datasets sorted by their average effect. In general, most models agree with the ground truth of the generated data, however, a few intriguing cases deviate from it. Firstly, CoxNAM on the Lin dataset, illustrated in Figure~\ref{fig:SimStudyLinearPH_CoxNAM_CoxNAM_WX}, ranked the features incorrectly, although it achieved the same performance as the other models. This could suggest that modeling the output as an additive function, as in the NAM models, does not guarantee providing correct explanations. Moreover, it indicates the importance of the role of the regularization terms in the CoxSE and CoxSENAM models. It also signifies the proposal of the hybrid model CoxSENAM as a better-regularized version of the NAM-based models. Another case is the CoxNAM explanations on the NonLinX dataset (Figure~\ref{fig:SimNonLinearPH3_3_CoxNAM_CoxNAM_WX}), where, although it ranked the features correctly, their relative magnitudes are not comparable to the ground truth in contrast to the same-performing model CoxSENAM (Figure~\ref{fig:SimNonLinearPH3_3_CoxSENAM_CoxSENAM_W}). Finally, although the CPH model ranked the features correctly on the NonLin dataset and incorrectly on the NonLinX dataset, both explanations can be considered random due to the model's poor random performance ($ \approx50\%$ C-index) on these two non-linear datasets.

\subsection{Feature Interactions}
\begin{figure}[H]
    \centering
    \includegraphics[width=0.7\linewidth]{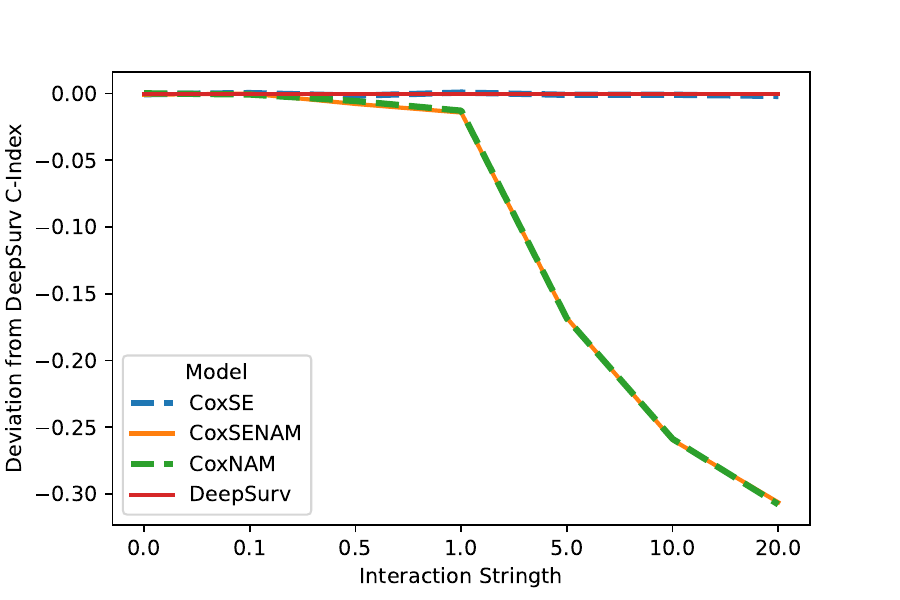}
    \caption{Impact of feature interactions on model performance - as a deviation from the DeepSurv performance}
    \label{fig:interaction}
\end{figure}

In the previous section, the results on the NonLinX dataset showed a large discrepancy between Deepsurv and CoxSE on one side and the NAM-based models on the other side. The two NAM-based models, as they have a separate model for each feature, learn the contribution of each feature based only on its corresponding feature, neglecting the interactions between the features. This is evident from the experimental results shown in Figure~\ref{fig:interaction}. In this experiment, we created seven synthetic datasets based on the log-risk function in Eq.~\ref{eq:interaction_dataset_equation} by varying the interaction strength factor $\lambda$.

\begin{equation}
    f(\bm{x})=3 x_0^2 + 2 x_1^2 + x_2^2 + \lambda x_0 x_1
    \label{eq:interaction_dataset_equation}
\end{equation}

DeepSurv model is the most flexible among the Cox-based models considered, as it imposes no structural assumptions on the function $f(\bm{x})$, beyond being represented by a feed-forward neural network. In contrast, CPH, CoxSE, CoxNAM, and CoxSENAM force structural constraints, making their hypothesis spaces subspaces of DeepSurv's. While this flexibility allows DeepSurv to approximate more complex relationships, its performance is still influenced by several factors, including the amount and quality of available data, the effectiveness of regularization, and optimization challenges during training. As a result, superior performance is not guaranteed in every setting. Figure~\ref{fig:interaction} shows the deviation of the models' performance from DeepSurv performance. The NAM-based models' performance degrades with the increase of the interaction strength, while the CoxSE model maintains an equivalent performance to DeepSurv. 
%

\subsection{Stability of Explanations}
The stability is assessed using the Lipschitz constant, Eq.~\ref{eq:lipschitz}, which quantifies the model's sensitivity to a small change in the input~\cite{Alvarez2018}. It estimates the maximum change in explanation for a small change in the vicinity of the point of interest.

\begin{equation}
    L = argmax_{\bm{x}_j \in \epsilon(\bm{x}_i)} \frac{\|f_{expl(\bm{x}_i)}-f_{expl(\bm{x}_j)}\|_2}{\|\bm{x}_i-\bm{x}_j\|_2}
    \label{eq:lipschitz}
\end{equation}
where $\epsilon(\bm{x}_i)$ is the neighborhood of $\bm{x}_i$.

We use the sample-based Lipschitz estimation in Equation~\ref{eq:lipschitz}. In this regard, the $\epsilon(x_i)$ are the set of points in the neighborhood of $x_i$. Moreover, we used the angle between the explanations as a distance metric. This choice is due to the special nature of Cox Proportional Hazards models. In such models, the scale of the output values does not matter as long as they maintain the same order. Consequently, two models that are the same except for a scaling factor result in different stability values.

\begin{figure}[H]
\centering
\begin{subfigure}{0.33\linewidth}
  \centering
  \includegraphics[width=1\linewidth]{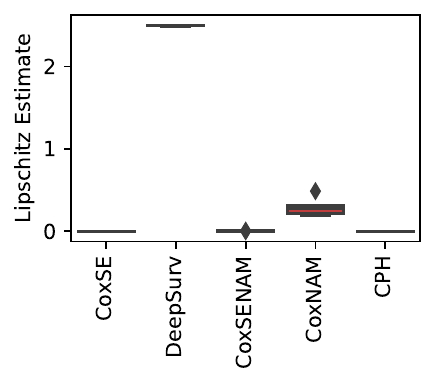}
  \caption{Lin Dataset }
  \label{fig:stability_Lin}
\end{subfigure}%
~
\begin{subfigure}{0.33\linewidth}
  \centering
  \includegraphics[width=1\linewidth]{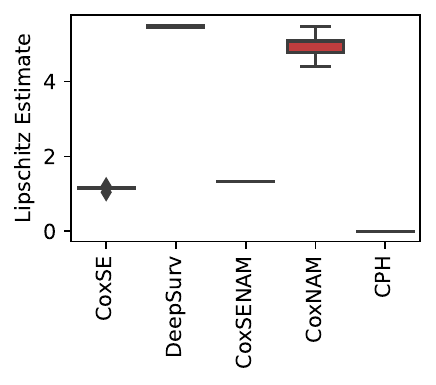}
  \caption{NonLin Dataset}
  \label{fig:stability_NonLin}
\end{subfigure}%
~
\begin{subfigure}{0.33\linewidth}
  \centering
  \includegraphics[width=1\linewidth]{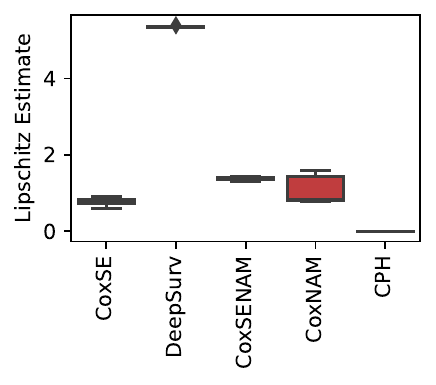}
  \caption{NonLinX Dataset}
  \label{fig:stability_NonLinX}
\end{subfigure}%

\begin{subfigure}{0.33\linewidth}
  \centering
  \includegraphics[width=1\linewidth]{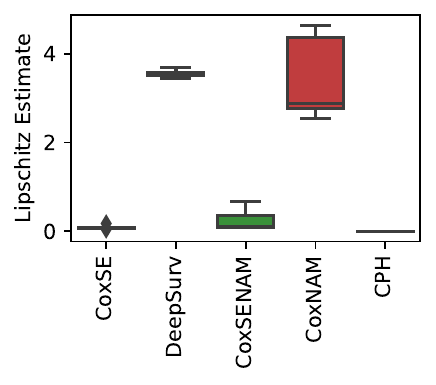}
  \caption{FLCHAIN Dataset}
  \label{fig:stability_Flchain}
\end{subfigure}%
~
\begin{subfigure}{0.33\linewidth}
  \centering
  \includegraphics[width=1\linewidth]{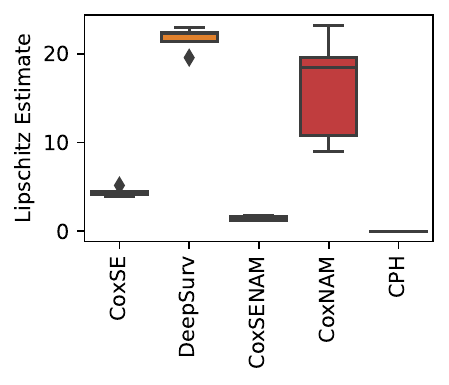}
  \caption{SEER Dataset}
  \label{fig:stability_SEER}
\end{subfigure}
\caption{Comparison of model stability across datasets, where lower Lipschitz estimates indicate greater stability}
\label{fig:stability}
\end{figure}

The results illustrated in Figure~\ref{fig:stability} show that the CPH model has the best stability. This is unsurprising as it is a linear model that has a global explanation for all points. Additionally, CoxSE and CoxSENAM models achieved better stability compared to CoxNAM and DeepSurv(+SHAP). This improvement can be attributed to the regularization term $\mathcal{L}_2$ weighted by $\alpha$ in their loss functions. In CoxSE and CoxSENAM models, the $\mathcal{L}_2$ term controls the model's flexibility locally. Higher $\alpha$ values result in less flexible models. In the limit, they converge to linear models for very high $\alpha$ values, i.e., the least flexibility and best stability.

\begin{figure}[H]
\centering
\begin{subfigure}{0.33\linewidth}
  \centering
  \includegraphics[width=1\linewidth]{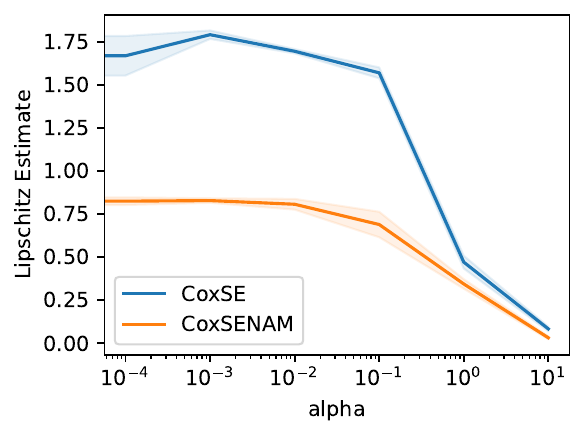}
  \caption{Stability vs. alpha }
  \label{fig:stability_vs_alpha}
\end{subfigure}%
~
\begin{subfigure}{0.33\linewidth}
  \centering
  \includegraphics[width=1\linewidth]{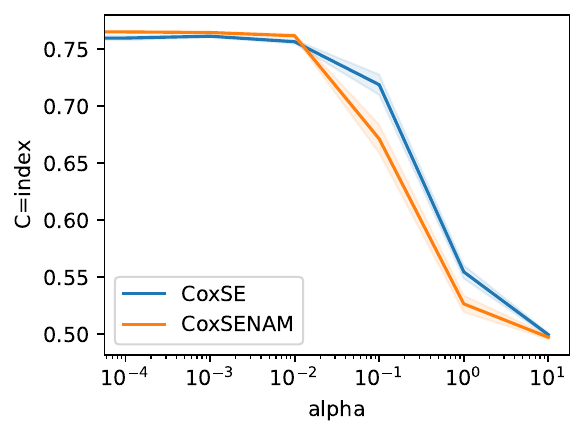}
  \caption{C-index vs. alpha}
  \label{fig:ci_vs_alpha}
\end{subfigure}%
~
\begin{subfigure}{0.33\linewidth}
  \centering
  \includegraphics[width=1\linewidth]{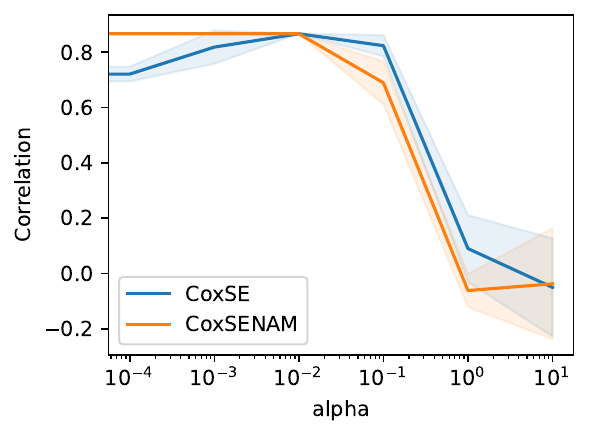}
  \caption{Correlation vs. alpha}
  \label{fig:corr_vs_alpha}
\end{subfigure}%

\begin{subfigure}{0.33\linewidth}
  \centering
  \includegraphics[width=1\linewidth]{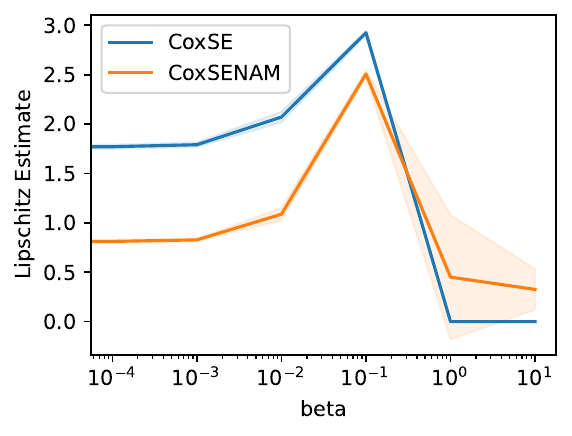}
  \caption{Stability vs. beta}
  \label{fig:stability_vs_beta}
\end{subfigure}%
~
\begin{subfigure}{0.33\linewidth}
  \centering
  \includegraphics[width=1\linewidth]{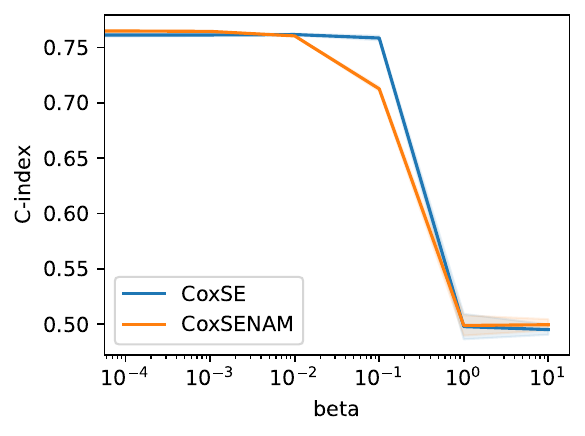}
  \caption{C-index vs. beta}
  \label{fig:ci_vs_beta}
\end{subfigure}%
~
\begin{subfigure}{0.33\linewidth}
  \centering
  \includegraphics[width=1\linewidth]{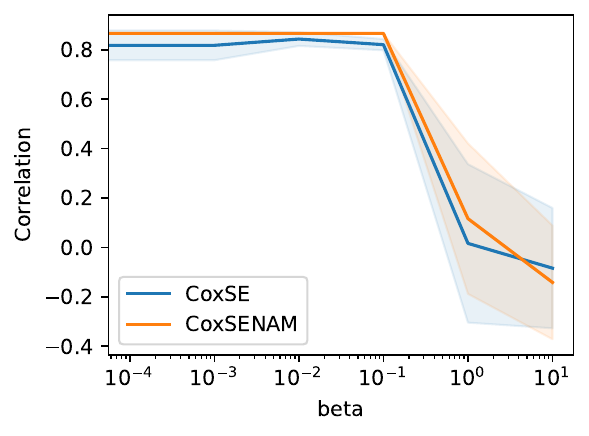}
  \caption{Correlation vs. beta}
  \label{fig:corr_vs_beta}
\end{subfigure}
\caption{The effect of $\alpha$ and $\beta$ on the stability, performance, and rank correlation with ground truth explanations}
\label{fig:stability_ci_corr_vs_alpha_beta}
\end{figure}

To dig deeper into the effect of the two hyperparameters $\alpha$ and $\beta$ on the stability and performance, we created synthetic datasets with non-linear log-risk as in Eq.~\ref{eq:interaction_dataset_equation} without interaction ($\lambda = 0$ ). Consequently, we trained CoxSE and CoxSENAM models varying $\alpha$ and $\beta$. The results in Figure~\ref{fig:stability_ci_corr_vs_alpha_beta} show that increasing $\alpha$ increases the stability as shown by the decline of the Lipschitz estimate in Figure~\ref{fig:stability_vs_alpha}. However, very high values of $\alpha$ limit the flexibility of the model, which causes a decline in the performance as shown in Figure~\ref{fig:ci_vs_alpha}. More importantly, for the CoxSE model, increasing $\alpha$ to a certain degree increased the agreement between the ground truth explanations and the predicted ones, as shown in Figure~\ref{fig:corr_vs_alpha}. However, after a certain degree, as the performance drops, the agreement with the ground truth explanations also drops due to underfitting.

On the other hand, increasing $\beta$ negatively affects the stability, Figure~\ref{fig:stability_vs_beta}, which can be attributed to the rigidity induced by the $\mathcal{L}_3$ term in the loss function that favors more sparse learned weights. Moreover, very high values of $\beta$ push all the learned weights towards zero, enhancing the stability; however, this causes a collapse in the model's performance and the correlation with the ground truth explanations, as illustrated in Figures~\ref{fig:stability_vs_beta},~\ref{fig:ci_vs_beta}, and~\ref{fig:corr_vs_beta}.

As demonstrated in the previous analysis, varying $\alpha$ and $\beta$ impacts both the model's stability, predictive performance, and alignment with ground truth explanations. Building on these findings, the analysis suggests that optimal values for $\alpha$ and $\beta$ can be selected by tuning the model to achieve peak performance, while choosing the highest possible regularization strengths that do not compromise that performance.

\subsection{Robustness to Non-informative Features}
The robustness is computed as the ratio of the informative to the non-informative contributions of the features. The higher the ratio, the more robust the model is to noisy features, in which case more contribution comes from informative features relative to non-informative features.  
\begin{eqnarray}
    \textit{Robustness} &=& \frac{1}{n} \sum_{i=0}^{n-1} \frac{y_{i}(\Omega)}{y_{i}(\overline{\Omega})} \\ \textit{where: }\nonumber \\
    y(\Psi) &=& \sum_{j \in \Psi} |C_j(\bm{x})|
\end{eqnarray}
where $\Psi$ represents a subset of the features, $\Omega$ is the set of informative features of $\bm{x}$, $\overline{\Omega}$ is its complement, i.e., the set of non-informative features, and $n$ the number of studied subjects. $C_j(\bm{x})$ is the contribution of the feature $x_j$ in the in the total output $y$, where it is ($C_j(\bm{x})=w_j(\bm{x})x_j$) for CoxSE, ($C_j(\bm{x})=w_j(x_j)x_j$) for CoxSENAM, and ($C_j(\bm{x})=f_j(x_j)$) for CoxNAM.

In this experiment, to guarantee that the three models, CoxSE, CoxNAM, and CoxSENAM, can model the data with comparable performances, we created synthetic datasets with non-linear log-risk as in Eq.~\ref{eq:interaction_dataset_equation} without interaction ($\lambda = 0$ ), varying the dataset size from $1$k to $50$k samples. Each dataset contains $10$ features, two of which are informative and the rest are non-informative features. To study the models' robustness and the effect of dataset size and the regularization parameters, we ran several experiments varying the dataset size. We also varied $\alpha$ and $\beta$ for the CoxSE and the CoxSENAM models. 
\begin{figure}[H]
\centering
\begin{subfigure}{0.4\columnwidth}
  \centering
  \includegraphics[width=1\linewidth]{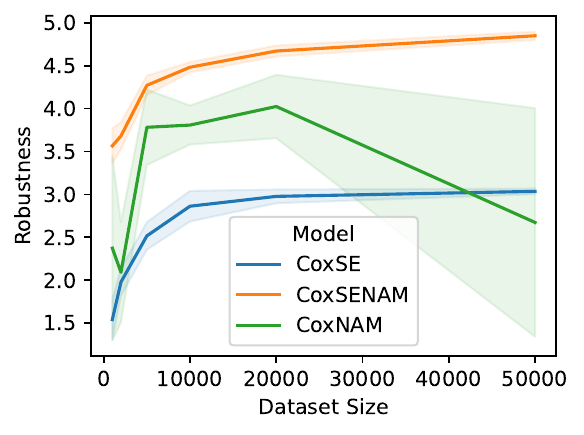}
  \caption{Robustness}
  \label{fig:n-vs-robustness-across-models}
\end{subfigure}%
~
\begin{subfigure}{0.4\columnwidth}
  \centering
  \includegraphics[width=1\linewidth]{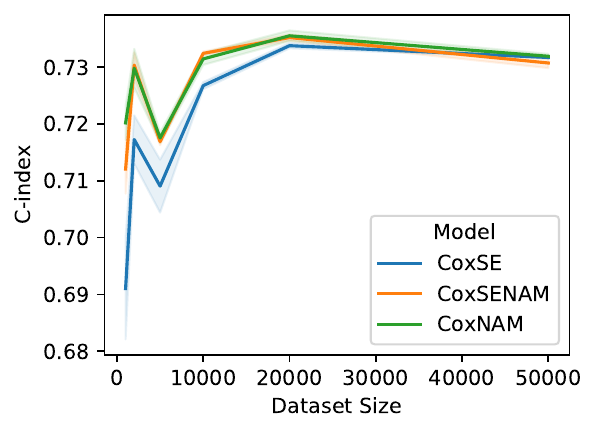}
  \caption{C-Index}
  \label{fig:n-vs-CI-across-models}
\end{subfigure}
\caption{Comparison between the models' performance and robustness across different dataset sizes}
\label{fig:robustness_ci_models}
\end{figure}

\begin{figure}[H]
\centering

\begin{subfigure}{.35\linewidth}
  \centering
  \includegraphics[width=1\linewidth]{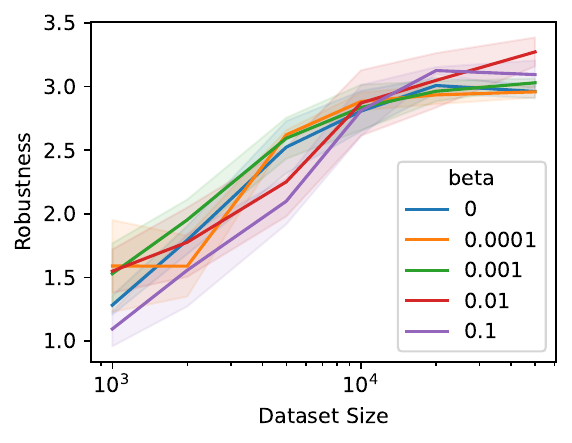}
  \caption{CoxSE: Robustness}
  \label{fig:CoxSE-alpha-0.001-n-vs-robustness-across-beta}
\end{subfigure}%
~
\begin{subfigure}{.35\linewidth}
  \centering
  \includegraphics[width=1\linewidth]{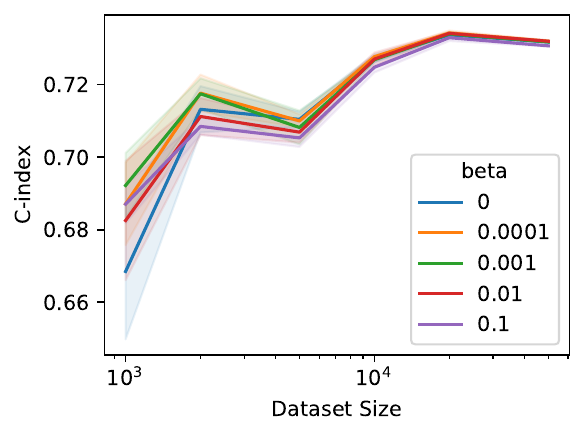}
  \caption{CoxSE: C-index}
  \label{fig:CoxSE-alpha-0.001-n-vs-CI-across-beta}
\end{subfigure}

\begin{subfigure}{.35\linewidth}
  \centering
  \includegraphics[width=1\linewidth]{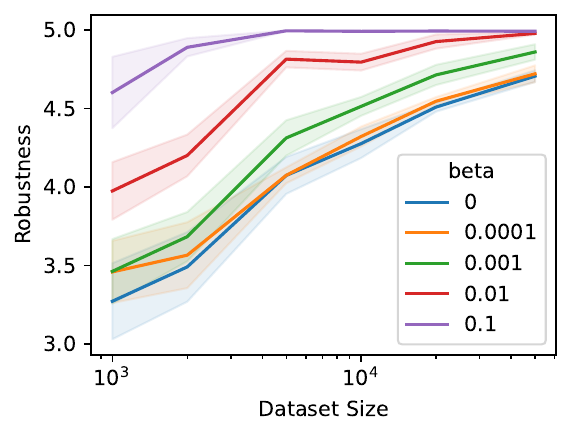}
  \caption{CoxSENAM: Robustness}
  \label{fig:CoxSENAM-alpha-0.001-n-vs-robustness-across-beta}
\end{subfigure}%
~
\begin{subfigure}{.35\linewidth}
  \centering
  \includegraphics[width=1\linewidth]{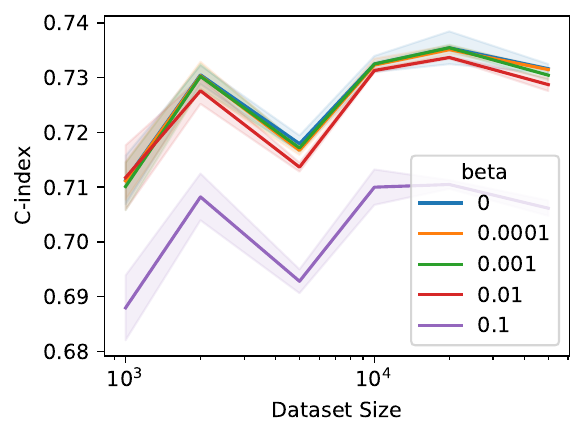}
  \caption{CoxSENAM: C-index}
  \label{fig:CoxSENAM-alpha-0.001-n-vs-CI-across-beta}
\end{subfigure}
\caption{Comparison between the models' performance and robustness across different dataset sizes and different $\beta$ values}
\label{fig:robustness_ci_beta}
\end{figure}

The results in Figure~\ref{fig:robustness_ci_models} show that all models benefit from increasing the dataset size in improving the C-index (Figure~\ref{fig:n-vs-CI-across-models}) and models' robustness (Figure~\ref{fig:n-vs-robustness-across-models}). Moreover, for very small and very large dataset sizes, CoxSE and CoxNAM have a comparable robustness score. At the same time, CoxNAM shows better robustness than CoxSE in med-size datasets as illustrated in Figure~\ref{fig:n-vs-robustness-across-models}. This can be attributed to the design structure of CoxNAM, where non-informative features only contribute to the final decision in the neural network in the form of additive noise. In CoxSE, however, computing feature relevances incorporates all the informative and non-informative features passed through the neural network in a complicated way. This makes it harder for the CoxSE model to isolate the noise. However, the hybrid model, CoxSENAM demonstrated better robustness, which can be attributed to the extra regularization term $\mathcal{L}_3$ weighted by $\beta$ in its loss function. To illustrate this, Figure~\ref{fig:robustness_ci_beta} shows the effect of dataset size and the change in $\beta$, the weight of the $\mathcal{L}_3$ term common for both CoxSE and CoxSENAM loss functions. Increasing $\beta$ does not seem to have any effect on the CoxSE model's robustness, however, the effect is significant in the case of CoxSENAM, as shown in Figures~\ref{fig:CoxSE-alpha-0.001-n-vs-robustness-across-beta} and \ref{fig:CoxSENAM-alpha-0.001-n-vs-robustness-across-beta} respectively.

\begin{figure}[H]
\centering
\begin{subfigure}{.35\linewidth}
  \centering
  \includegraphics[width=1\linewidth]{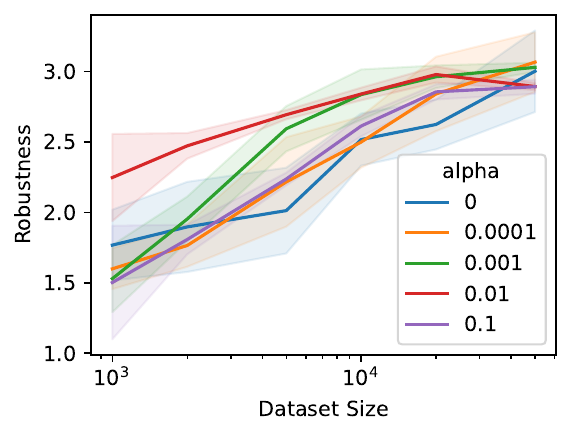}
  \caption{CoxSE: Robustness}
  \label{fig:CoxSE-beta-0.001-n-vs-robustness-across-alpha}
\end{subfigure}%
~
\begin{subfigure}{.35\linewidth}
  \centering
  \includegraphics[width=1\linewidth]{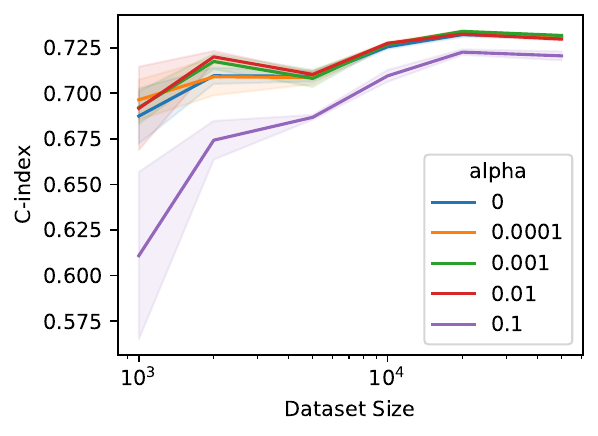}
  \caption{CoxSE: C-index}
  \label{fig:CoxSE-beta-0.001-n-vs-CI-across-alpha}
\end{subfigure}

\begin{subfigure}{.35\linewidth}
  \centering
  \includegraphics[width=1\linewidth]{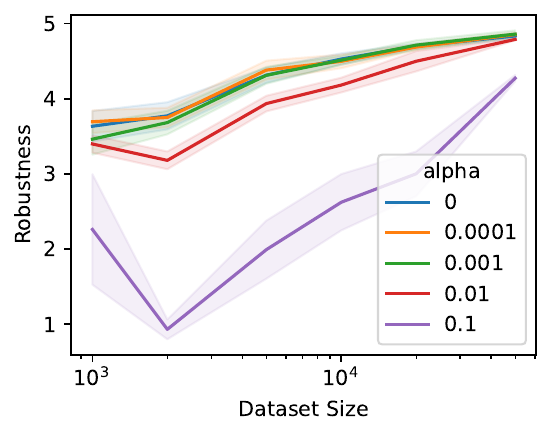}
  \caption{CoxSENAM: Robustness}
  \label{fig:CoxSENAM-beta-0.001-n-vs-robustness-across-alpha}
\end{subfigure}%
~
\begin{subfigure}{.35\linewidth}
  \centering
  \includegraphics[width=1\linewidth]{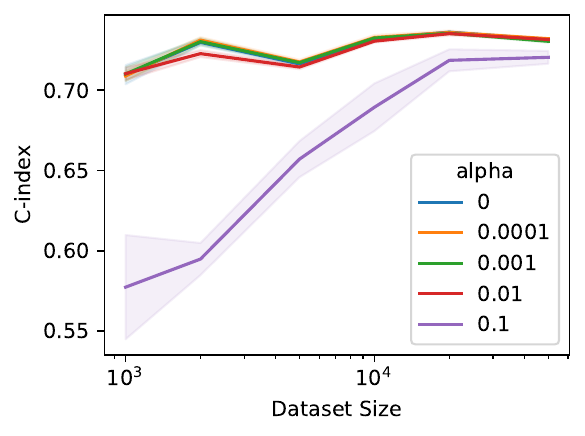}
  \caption{CoxSENAM: C-index}
  \label{fig:CoxSENAM-beta-0.001-n-vs-CI-across-alpha}
\end{subfigure}%
\caption{Comparison between the models' performance and robustness across different dataset sizes and different $\alpha$ values}
\label{fig:robustness_ci_alpha}
\end{figure}

On the other hand, increasing $\alpha$ doesn't affect the robustness of CoxSE and CoxSENAM models. However, for very high values of $\alpha$, the models' performance drops significantly due to underfitting, which in turn leads to a decline in the CoxSENAM robustness, as illustrated in Figure~\ref{fig:robustness_ci_alpha}.

\section{Explanations Assessment on Real Datasets}

In this section, we discuss the results of the two real datasets, FLCHAIN and SEER. With the existence of collinearities and correlations between the features, models can rely on different subsets of features, which result in different explanations. However, as we do not know the ground truth of these datasets, an interesting validation method would be to compare how consistent the explanations provided by the model are with the explanations provided by an external method like SHAP applied to the model itself.

The SHAP method computes the features' contribution to the model's decision. Whereas, CoxSE and CoxSENAM models produce explanations in terms of locally linear weights, which represent the sensitivity to change in the respective features. Based on such weights, we can compute the contribution in a similar way to how SHAP does for a linear model. 

Given a linear model trained on a dataset with $m$ features and has weights $\beta_i : i \in 0..m-1$, the Shapley Value $\Phi_i$ of the feature $x_i$ weighted by $\beta_i$ can be written as:
\begin{equation}
    \Phi_i = \beta_i (x_i - \mathbb{E}(x_i))
    \label{eq:shapley_value}
\end{equation}
Based on Eq.~\ref{eq:shapley_value}, we compute the contributions of the features for the CoxSE and the CoxSENAM by replacing the constant weights with their variable weights as:
\begin{equation}
    \Phi_i = w_i(\bm{x}) (x_i - \mathbb{E}(x_i))
    \label{eq:shapley_value_se}
\end{equation}

We compared the rank order of the aggregated feature importance provided by each model against the rank order of the aggregated SHAP explanation of the same model. We used the weighted-$\tau$~\cite{Vigna2015}, which is a weighted version of Kendall-$\tau$~\cite{Kendall1938} rank correlation index. The weighted-$\tau$ penalizes the disagreement in the higher-ranked items more than the lower ones. This makes it convenient for comparing feature rankings. 
\begin{figure}[H]
\centering
\begin{subfigure}{0.4\linewidth}
  \centering
  \includegraphics[width=1\linewidth]{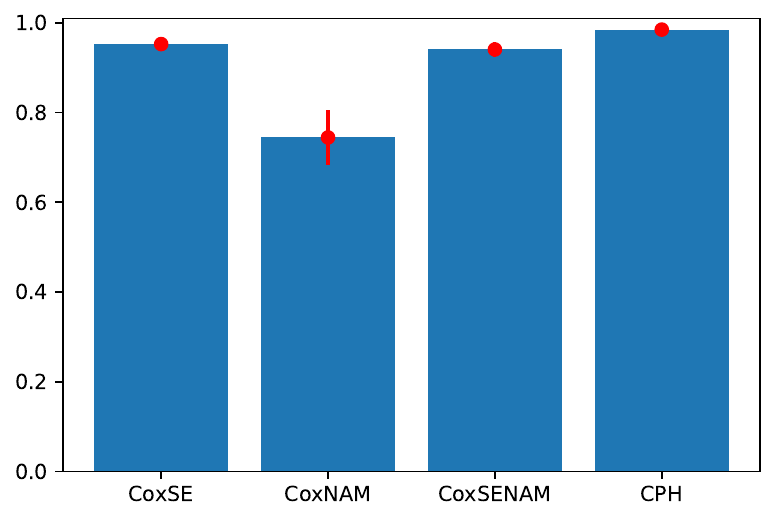}
  \caption{FLCHAIN}
  \label{fig:consistncy_flchain_wo_deepsurv}
\end{subfigure}%
~
\begin{subfigure}{0.4\linewidth}
  \centering
  \includegraphics[width=1\linewidth]{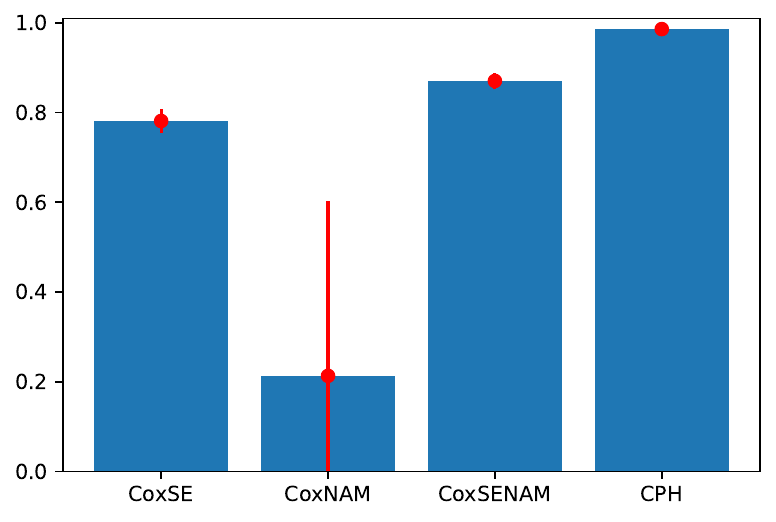}
  \caption{SEER}
  \label{fig:consistncy_SEERBreastCancer_wo_deepsurv}
\end{subfigure}
\caption{Rank correlation between the models' explanations and their own Shapley Values}
\label{fig:consistncy_models}
\end{figure}
Except for the approximation error of the SHAP method, the results in Figure~\ref{fig:consistncy_models} show almost a perfect correlation between the linear CPH model's explanations and its Shapley Values. This is understandable due to the direct relationship between Shapley Values and the linear weights as shown in Eq.~\ref{eq:shapley_value}.
The results also show a clear advantage of the SENN-based models (CoxSE and CoxSENAM), which have much higher correlations with their own Shapley Values than the CoxNAM model. This can be attributed to the local linearity of the models encouraged by the regularization term $\mathcal{L}_2$. 

It is worth noting that, on the synthetic datasets, there was a perfect match between the ranking of the aggregated explanations of all models with the ranking of their own aggregated Shapley Values, except for the case of CoxNAM on the Lin dataset shown in Figure~\ref{fig:SimStudyLinearPH_CoxNAM_CoxNAM_WX}. This can explain the lower correlation between CoxNAM and its SHAP explanations in the case of the real datasets. Moreover, this strengthens the hypothesis that transforming the function into an additive function does not guarantee valid explanations and highlights the importance of the regularization terms in the SENN-based models in the validity of the generated explanations.

\subsection{Case Study: SEER Breast Cancer Dataset}
\begin{figure}[H]
\centering
\begin{subfigure}{0.5\linewidth}
  \centering
  \includegraphics[width=1\linewidth]{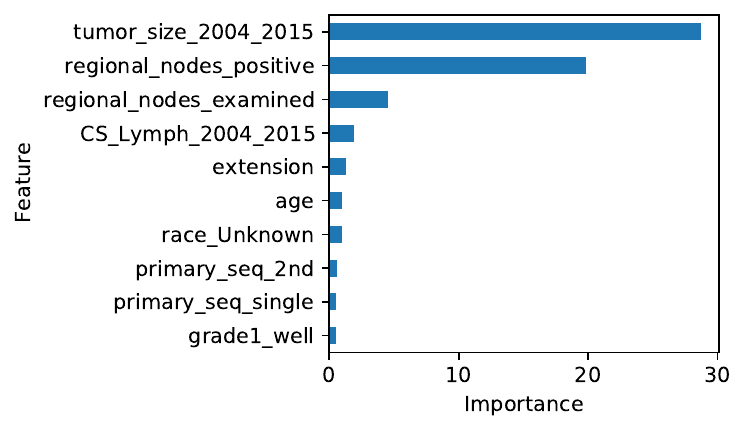}
  \caption{CoxSE}
  \label{fig:coxse_seer_explanations}
\end{subfigure}%
~
\begin{subfigure}{0.5\linewidth}
  \centering
  \includegraphics[width=1\linewidth]{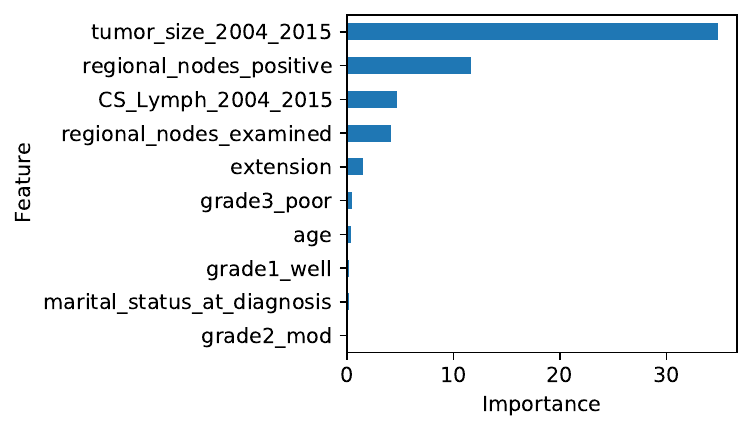}
  \caption{CoxSENAM}
  \label{fig:coxsenam_seer_explanations}
\end{subfigure}
\caption{The top 10 important features of the SEER dataset identified by CoxSE and CoxSENAM models}
\label{fig:coxse_coxsenam_seer_explanations}
\end{figure}

To demonstrate the practical utility of our proposed models, we present a case study using the SEER breast cancer dataset. This dataset contains comprehensive clinical features from a large cohort of breast cancer patients, including tumor size, lymph node involvement, extension, and other pathological variables. We applied both CoxSE and CoxNAMSE models to analyze feature relevance across the patient population. Both proposed models consistently identified the same top five prognostic factors: tumor size, number of lymph nodes containing tumor, number of lymph nodes removed and examined, tumor extension, and the highest specific lymph node chain involved, as illustrated in Figure~\ref{fig:coxse_coxsenam_seer_explanations}. 
These variables correspond closely to the core anatomical and pathological components of the TNM classification system, where tumor size and extension define the T category and lymph node involvement defines the N category~\cite{giuliano_eighth_2018}, reflect the prognostic emphasis found in Nottingham Prognostic Index (NPI)~\cite{galea_nottingham_1992}, and are consistent with insights reported in the literature~\cite{el_saghir_effects_2006}. This consistency lends support to the clinical plausibility of the model-generated explanations and suggests their potential utility in supporting clinical understanding and interpretation of survival outcomes, particularly in settings where transparency and interpretability are important. When compared with the baseline approaches (Figure 15), including CoxNAM, DeepSurv + SHAP, CPH, and RSF, the proposed models showed a noticeably different pattern in how they explained predictions. While the baseline models often ranked some demographic factors (such as age and race) and tumor grade among the top five features, the proposed models focused mainly on TNM-related variables. Moreover, the proposed methods exhibited greater differentiation in feature importance magnitudes, showing a clearer separation between dominant and secondary predictors than observed in other methods.

\begin{figure}[H]
\centering
\begin{subfigure}{0.5\linewidth}
  \centering
  \includegraphics[width=1\linewidth]{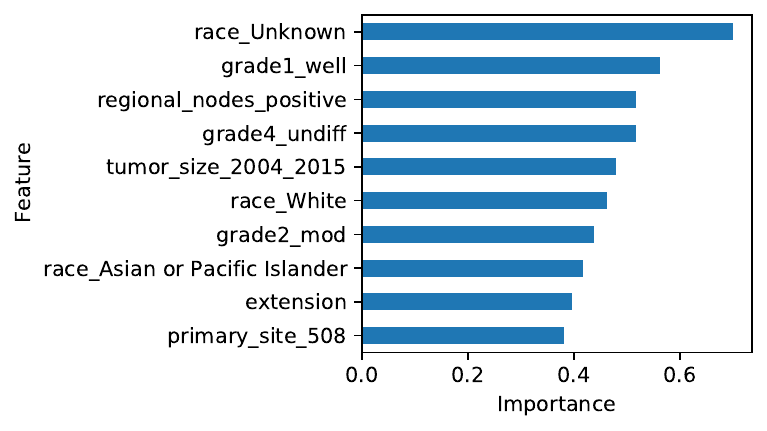}
  \caption{CoxNAM}
  \label{fig:coxnam_seer_explanations}
\end{subfigure}%
~
\begin{subfigure}{0.5\linewidth}
  \centering
  \includegraphics[width=1\linewidth]{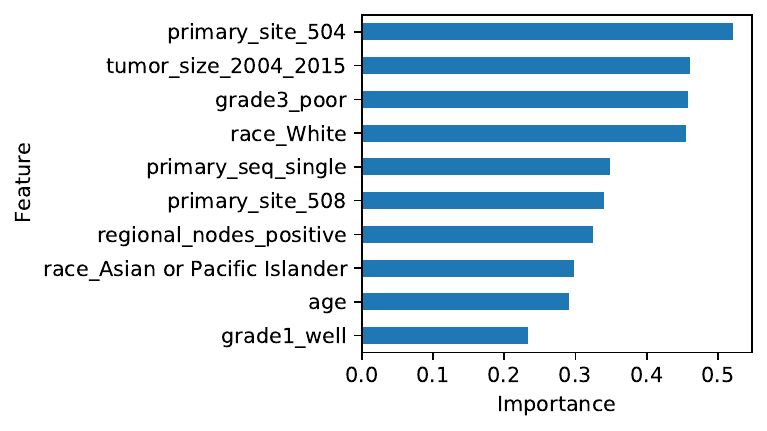}
  \caption{DeepSurv + SHAP}
  \label{fig:deepsurv_seer_explanations}
\end{subfigure}

\begin{subfigure}{0.5\linewidth}
  \centering
  \includegraphics[width=1\linewidth]{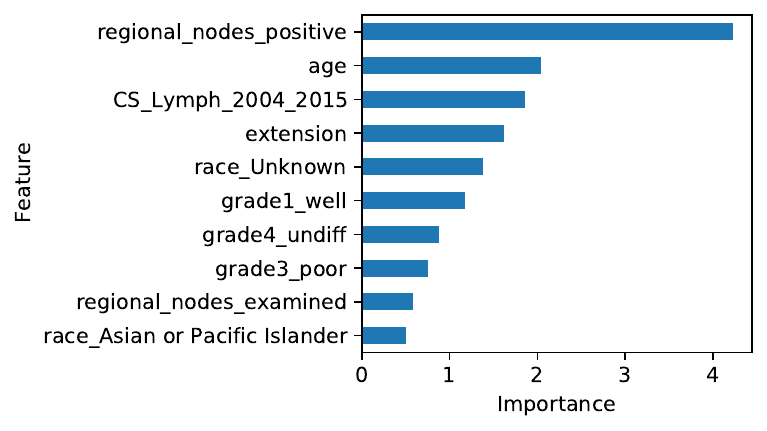}
  \caption{CPH}
  \label{fig:cph_seer_explanations}
\end{subfigure}%
~
\begin{subfigure}{0.5\linewidth}
  \centering
  \includegraphics[width=1\linewidth]{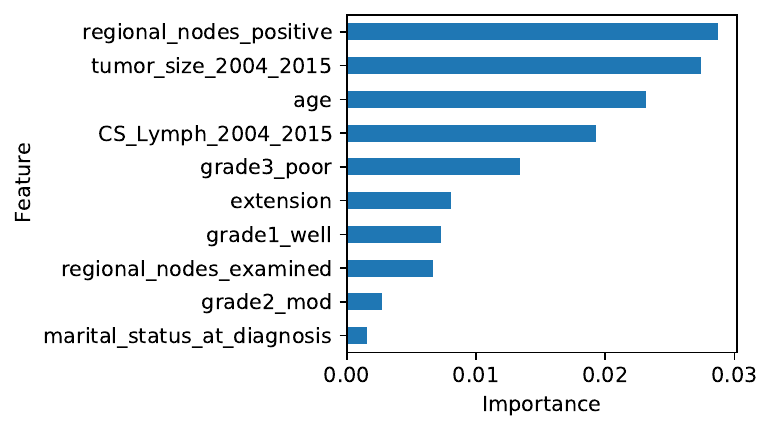}
  \caption{RSF}
  \label{fig:rsf_seer_explanations}
\end{subfigure}

\caption{The top 10 important features of the SEER dataset identified by CoxNAM, DeepSurv, CPH, and RSF models}
\label{fig:others_seer_explanations}
\end{figure}

\section{Conclusion}
In this work, we explored how self-explaining neural networks with the CPH model can be adapted for survival modeling. We proposed two CPH-based models, CoxSE relying on the SENN structure, and CoxSENAM, which is a hybrid model between the SENN and the NAM structures. The two models share the same type of output and loss function, providing local explanations as a locally linear approximation of the underlying function. In this regard, unlike Shapley Values and the CoxNAM model, which provide explanations as feature contributions, our models' explanations come in the form of local weights. As such, they highlight the sensitivity of the outcome to changes in the features in the vicinity of the point of interest. However, we illustrated that feature contributions can be computed directly from such weights, similar to the case of a linear model, however, with variable weights.     

We performed several experiments with synthetic and real datasets, which highlighted the properties of the proposed models. In terms of performance, CoxSE was shown to be a flexible model that matches the performance of its black-box counterpart, DeepSurv. However, similar to the CoxNAM model, CoxSENAM is less flexible due to the NAM structure, which cannot handle interactions between the features. Nevertheless, CoxSE and CoxSENAM showed better stability than CoxNAM and DeepSurv (explained by SHAP) and better matching with the ground truth than CoxNAM. This is due to the regularization term weighted by $\alpha$ in their loss function that directly encourages local similarity of explanations. 

On the other hand, the NAM-based models showed better robustness to noisy features than the CoxSE model. This can be attributed to the structure of the NAM models, where the noise contribution to the output comes as additive noise. This makes it easier for the NAM-based models to isolate the noise, transforming it into a constant value added to the output. However, our hybrid model, CoxSENAM, showed better robustness than CoxNAM, assigning less importance to noisy features. This can be attributed to the second regularization term, weighted by $\beta$, encouraging the noise contribution to be close to zero.

Lastly, using real datasets, our models, CoxSE and CoxSENAM, demonstrated better consistency between their intrinsic explanations and the post-hoc explanation provided by SHAP. This was also observed on the synthetic datasets, which arguably indicates the validity of such explanations in describing the underlying contributions of the features. On the other hand, CoxNAM demonstrated less consistency with its post-hoc explanations on both the real and synthetic datasets.

This work investigated how to employ the SENN model in Survival Analysis, relying on the Cox Proportional Hazards assumptions and using the partial log-likelihood objective function. This poses a restriction on the application of such a model to non-proportional hazards type of data. In non-proportional hazards scenarios, relying on models that assume proportionality can result in biased risk estimates and misleading interpretations of covariate effects over time. Consequently, the use of such models in these contexts may compromise both the validity of the predictions and the reliability of the derived explanations, as highlighted in \cite{Schemper92,Alabdallah2022}. However, other survival model formulations can be used, such as direct regression of survival time or estimating the Probability Mass Function (PMF) in a discrete-time manner, which we leave for future work. Moreover, in this work, we assumed the original feature space to be explainable, which is the most common form of survival data. Consequently, we discarded the interpretable feature extraction part of the network (the autoencoder) along with its reconstruction loss function. However, in some cases, where the input space is not interpretable, like the case of image input, the feature-extraction part can be brought back, which is out of the scope of this work.

\appendix
\section{Hyperparameters Tuning}
\label{appendix_a}
We performed hyperparameter tuning using five-fold cross-validation. The final hyperparameter configurations for all models are presented in Tables~\ref{tbl_hp_coxse}\-~\ref{tbl_hp_deephit}.

\begin{table}[H]
\caption{hyperparameters of the CoxSE model}
\centering
\begin{tabular}{l c c c c c} 
 \hline
 Hyperparameter & Lin & NonLin & NonLinX & Flchain & SEER \\ 
 \hline
 No. Layers     & 4    & 1     & 1    & 4    & 1      \\
 No. Nodes      & 128  & 4     & 32   & 4    & 4      \\ 
 L2             & 0    & 0.5   & 0.5  & 0.5  & 0.5    \\ 
 Dropout        & 0.4  & 0     & 0    & 0.3  & 0      \\
 Learning Rate  & 0.01 & 0.001 & 0.1  & 0.01 & 0.0009 \\
 alpha          & 0.5  & 3e-6  & 1e-5 & 4e-5 & 1e-5   \\
 beta           & 0    & 5e-4  & 1e-4 & 6e-3 & 3e-5   \\
 \hline
\end{tabular}
\label{tbl_hp_coxse}
\centering
\end{table}

\begin{table}[H]
\caption{hyperparameters of the CoxSENAM model}
\centering
\begin{tabular}{l c c c c c} 
 \hline
 Hyperparameter & Lin & NonLin & NonLinX & Flchain & SEER \\ 
 \hline
 No. Layers     & 1     & 1     & 1    & 4     & 4     \\
 No. Nodes      & 4     & 128   & 64   & 32    & 16    \\ 
 L2             & 0.5   & 1     & 0    & 0.001 & 0.5   \\ 
 Dropout        & 0     & 0.4   & 0.1  & 0.4   & 0     \\
 Learning Rate  & 0.1   & 0.006 & 0.1  & 0.003 & 0.001 \\
 alpha          & 0.5   & 9e-6  & 3e-3 & 6e-5  & 7e-5  \\
 beta           & 0.005 & 9e-6  & 5e-6 & 5e-5  & 5e-6  \\
 \hline
\end{tabular}
\label{tbl_hp_coxsenam}
\centering
\end{table}

\begin{table}[H]
\caption{hyperparameters of the CoxNAM model}
\centering
\begin{tabular}{l c c c c c} 
 \hline
 Hyperparameter & Lin & NonLin & NonLinX & Flchain & SEER \\ 
 \hline
 No. Layers     & 2   & 4      & 2    & 4     & 4      \\
 No. Nodes      & 32  & 128    & 128  & 64    & 16     \\ 
 L2             & 0.7 & 1      & 0.05 & 0     & 0.05   \\ 
 Dropout        & 0   & 0      & 0    & 0     & 0      \\
 Learning Rate  & 0.1 & 0.0009 & 0.05 & 0.001 & 0.0005 \\
 \hline
\end{tabular}
\label{tbl_hp_coxnam}
\centering
\end{table}

\begin{table}[H]
\caption{hyperparameters of the DeepSurv model}
\centering
\begin{tabular}{l c c c c c} 
 \hline
 Hyperparameter & Lin & NonLin & NonLinX & Flchain & SEER \\ 
 \hline
 No. Layers     & 3    & 1     & 1      & 1      & 1      \\
 No. Nodes      & 4    & 128   & 128    & 16     & 32     \\ 
 L2             & 0.1  & 0.01  & 0.5    & 1      & 1      \\ 
 Dropout        & 0    & 0     & 0      & 0.1    & 0.2    \\
 Learning Rate  & 0.05 & 0.001 & 0.0001 & 0.0005 & 0.0005 \\
 \hline
\end{tabular}
\label{tbl_hp_deepsurv}
\centering
\end{table}

\begin{table}[H]
\caption{hyperparameters of the CPH model}
\centering
\begin{tabular}{l c c c c c} 
 \hline
 Hyperparameter & Lin & NonLin & NonLinX & Flchain & SEER \\ 
 \hline
 Learning Rate  & 0.1 & 0.0003 & 0.03 & 0.09 & 0.01 \\
 \hline
\end{tabular}
\label{tbl_hp_cph}
\centering
\end{table}

\begin{table}[H]
\caption{hyperparameters of the RSF model}
\centering
\begin{tabular}{l c c c c c} 
 \hline
 Hyperparameter     & Lin    & NonLin& NonLinX & Flchain & SEER \\ 
 \hline
 $\texttt{min\_samples\_split}$  & 20      & 10    & 10    & 10   & 10 \\
 $\texttt{min\_samples\_leaf}$   & 5       & 15    & 15    & 15   & 20 \\
 $\texttt{max\_features}$       & log2    & sqrt  & sqrt  & sqrt & log2 \\ 
 $\texttt{max\_depth}$          & 10      & 10    & 10    & 10   & 20 \\
 \hline
\end{tabular}
\label{tbl_hp_rsf}
\centering
\end{table}

\begin{table}[H]
\caption{hyperparameters of the DeepHit model}
\centering
\begin{tabular}{l c c c c c} 
 \hline
 Hyperparameter & Lin    & NonLin& NonLinX & Flchain & SEER \\ 
 \hline
 No. bins       & 500    & 200    & 1000  & 1000   & 200 \\
 alpha          & 1e-1   & 1e-2   & 1e-2  & 5e-2   & 1e-1 \\ 
 sigma          & 9e-1   & 1.0    & 1e-3  & 9e-1   & 5e-1 \\ 
 Dropout        & 0.4    & 0      & 0.1   & 0.2    & 0.4 \\
 No. Layers     & 2      & 2      & 2     & 2      & 2 \\
 No. Nodes      & 32     & 128    & 16    & 64     & 64 \\
 Learning Rate  & 0.03   & 0.004 & 0.0009 & 0.0005 & 0.003 \\
 \hline
\end{tabular}
\label{tbl_hp_deephit}
\centering
\end{table}

\bibliographystyle{elsarticle-num}
\bibliography{main}

\end{document}